\documentclass[journal]{IEEEtran}

\usepackage{graphicx}

\usepackage{amsmath}
\usepackage{amssymb}
\usepackage{amsfonts}

\usepackage{xcolor}
\usepackage{adjustbox}
\usepackage{colortbl}
\usepackage{tabularx}

\usepackage{booktabs}
\usepackage{multirow}
\usepackage{subfig}
\usepackage{cite}
\usepackage{floatflt}
\usepackage{wrapfig}
\usepackage[linesnumbered, ruled,vlined]{algorithm2e}
\usepackage{moresize}
\usepackage{url}

\usepackage{tablefootnote}
\usepackage{bigfoot}
\usepackage{pdflscape}

\usepackage{hyperref}
\usepackage{authblk}

\usepackage{colortbl}
\usepackage{tabularx}

\def\vector#1{\mbox{\boldmath $#1$}}

\newcommand{\PreserveBackslash}[1]{\let\temp=\\#1\let\\=\temp}
\newcolumntype{C}[1]{>{\PreserveBackslash\centering}p{#1}}

\ifCLASSINFOpdf
\else
\fi
\begin{document}
%


\title{Benchmarking Feature-based Algorithm Selection Systems for Black-box Numerical Optimization}







%
%
%

\author{
Ryoji~Tanabe,~\IEEEmembership{Member,~IEEE}
\thanks{R. Tanabe is with Faculty of Environment and Information Sciences, Yokohama National University, Yokohama, Japan, and also with Center for Advanced Intelligence Project, RIKEN, Tokyo, Japan. e-mail: (rt.ryoji.tanabe@gmail.com).}
}

\maketitle


\begin{abstract}

Feature-based algorithm selection aims to automatically find the best one from a portfolio of optimization algorithms on an unseen problem based on its landscape features.
Feature-based algorithm selection has recently received attention in the research field of black-box numerical optimization.
However, there is still room for analysis of algorithm selection for black-box optimization.
Most previous studies have focused only on whether an algorithm selection system can outperform the single-best solver in a portfolio.
In addition, a benchmarking methodology for algorithm selection systems has not been well investigated in the literature.
In this context, this paper analyzes algorithm selection systems on the 24 noiseless black-box optimization benchmarking functions.
First, we demonstrate that the first successful performance measure is more reliable than the expected runtime measure for benchmarking algorithm selection systems.
Then, we examine the influence of randomness on the performance of algorithm selection systems.
We also show that the performance of algorithm selection systems can be significantly improved by using sequential least squares programming as a pre-solver.
We point out that the difficulty of outperforming the single-best solver depends on algorithm portfolios, cross-validation methods, and dimensions.
Finally, we demonstrate that the effectiveness of algorithm portfolios depends on various factors.
These findings provide fundamental insights for algorithm selection for black-box optimization.

\end{abstract}


%
\IEEEpeerreviewmaketitle

\section{Introduction}
\label{sec:introduction}

\IEEEPARstart{B}LACK-BOX numerical optimization aims to find a solution $\vector{x} \in \mathbb{R}^n$ with an objective value $f(\vector{x})$ as small as possible without any explicit knowledge of the objective function $f: \mathbb{R}^n \rightarrow \mathbb{R}$.
Here, $n$ is the dimension of the solution space.
This paper considers only single-objective noiseless black-box optimization.
A number of derivative-free black-box optimizers have been proposed in the literature, including mathematical optimization approaches, Bayesian optimization approaches, and evolutionary optimization approaches.
In general, the best optimizer depends on the characteristics of a given problem \cite{HansenARFP10}.
It is also difficult for a user to select an appropriate optimizer for her/his problem through a trial-and-error process.
Thus, automatic algorithm selection is essential for practical black-box optimization.


The algorithm selection problem \cite{Rice76,LindauerRK19} involves selecting the best one from a portfolio of $k$ algorithms $\mathcal{A} = \{a_1, ..., a_k\}$ on a set of problem instances $\mathcal{I}$ in terms of a performance measure $m: \mathcal{A} \times \mathcal{I} \rightarrow \mathbb{R}$.
%
The algorithm selection problem is a fundamental research topic in the fields of artificial intelligence and evolutionary computation \cite{Smith-Miles08,Kotthoff16,KerschkeHNT19}.

Feature-based offline algorithm selection is one of the most popular approaches for the algorithm selection problem \cite{Smith-Miles08,Kotthoff16,KerschkeHNT19}.
First, a feature-based algorithm selection approach computes numerical features of a given problem.
It is desirable that the features well capture the characteristics of the problem.
Then, the approach predicts the most promising algorithm $a^{\mathrm{best}}$ from a pre-defined portfolio $\mathcal{A}$ based on the features.
Machine learning techniques are generally employed to build a selection model.
Finally, $a^{\mathrm{best}}$ is applied to the problem.
Feature-based algorithm selection has demonstrated its effectiveness on a wide range of problem domains, including the propositional satisfiability problem (SAT) \cite{XuHHL08}, the traveling salesperson problem (TSP) \cite{KerschkeKBHT18}, answer set programming (ASP) \cite{HoosLS14}, and multi-objective optimization \cite{LiefoogheVLZM21}.

Six recent studies \cite{BischlMTP12,AbellMT13,DerbelLVAT19,KerschkeT19,JankovicPED21,MunozK21} have reported promising results of feature-based algorithm selection for black-box numerical optimization\footnote{We say that a previous study relates to algorithm selection only when it \textit{actually} performed algorithm selection. Although some previous studies identified ``performance prediction'' with ``algorithm selection'', we strictly distinguish the two different tasks.}.
All of them performed algorithm selection on the 24 noiseless BBOB functions \cite{hansen2012fun}.
Throughout this paper, we denote the noiseless BBOB functions as the BBOB functions.
Except for \cite{AbellMT13}, these studies used exploratory landscape analysis (ELA) \cite{MersmannBTPWR11} for feature computation, where ELA computes a set of numerical features of a given problem based on a set of solutions.
In addition to ELA, the study \cite{DerbelLVAT19} proposed a feature computation method based on the tree constructed by simultaneous optimistic optimization (SOO) \cite{Munos11}.
The study \cite{DerbelLVAT19} also investigated the effectiveness of the SOO-based features.
The results in the previous studies showed that feature-based algorithm selection systems could potentially outperform the single-best solver (SBS) on the BBOB function suite, where SBS is the best optimizer in a portfolio $\mathcal{A}$ across all function instances.
Here, we use the term ``an algorithm selection system'' to represent a whole system that includes, e.g., a feature computation method, an algorithm selection method, and an algorithm portfolio.

In other words, the previous studies have mainly focused only on whether an algorithm selection system can perform better than the SBS.
Although an algorithm selection system consists of many elements, their influence has not been investigated in the literature.
A better understanding of algorithm selection systems is needed for the next step.
More importantly, a benchmarking methodology has not been well standardized in the field of algorithm selection for black-box optimization.
Table \ref{suptab:prev_as_systems} in the supplementary file shows the experimental settings in the five previous studies \cite{BischlMTP12,DerbelLVAT19,KerschkeT19,JankovicPED21,MunozK21}, except for \cite{AbellMT13}.
We do not explain Table \ref{suptab:prev_as_systems} due to the paper length limitation, but the experimental settings in the five previous studies are different, including a cross-validation method and an algorithm portfolio.
Thus, none of the five previous studies \cite{BischlMTP12,DerbelLVAT19,KerschkeT19,JankovicPED21,MunozK21} adopted the same experimental setting.

\noindent \textbf{Contributions}

In this context, this paper analyzes feature-based offline algorithm selection systems for black-box numerical optimization.
Through a benchmarking study, this paper addresses the following six research questions.

\noindent \textbf{RQ1} \textit{Is the expected runtime reliable for benchmarking algorithm selection systems?}
The expected runtime (ERT) \cite{AugerH05a} is a general performance measure for benchmarking black-box optimizers in the fixed-target scenario \cite{HansenABTT16}.
Most previous studies also evaluated the performance of algorithm selection systems by using the ERT.
However, as discussed in \cite{DerbelLVAT19}, the ERT is sensitive to the maximum number of function evaluations for an unsuccessful run.
When different optimizers in a portfolio use different termination conditions, the ERT may incorrectly evaluate the performance of algorithm selection systems.

\noindent \textbf{RQ2} \textit{How does the performance of algorithm selection systems depend on randomness?}
Most operations in algorithm selection systems include randomness, e.g., the generation of the sample and the computation of features.
However, the previous studies for black-box optimization \cite{BischlMTP12,AbellMT13,DerbelLVAT19,KerschkeT19,JankovicPED21,MunozK21} performed only a single run of an algorithm selection system.
In \cite{LindauerRK19}, Lindauer et al. demonstrated that the performance of the winner of the algorithm selection competition in 2017 significantly depends on a random seed.
Thus, it is necessary to understand the influence of randomness on the performance of algorithm selection systems for black-box optimization.
%
The previous studies \cite{RenauDDD20,MunozK21} focused on randomness in the sampling phase.
The study \cite{CameronHL16} pointed out that the VBS performance of a portfolio can be overestimated when the portfolio includes randomized solvers.
The study \cite{HurleyO15} investigated the influence of the runtime variation of randomized SAT solvers on the accuracy of runtime predictors.
In contrast, we are interested in randomness in the whole process of algorithm selection.



\noindent \textbf{RQ3} \textit{How much can a pre-solver improve the performance of an algorithm selection system? Which optimizer is suitable for a pre-solver?}
Some modern algorithm selection systems in the field of artificial intelligence (e.g., \textsc{SATzilla} \cite{XuHHL08} and 3S \cite{KadiogluMSSS11}) adopt the concept of pre-solving.
Here, pre-solving is an approach that aims to solve easy problem instances quickly before the algorithm selection process starts.
In contrast, no previous study has used a pre-solver for black-box optimization. 
As investigated in \cite{KerschkeT19}, algorithm selection systems generally perform poorly on easy function instances, e.g., $f_1$ (the Sphere function) in the BBOB function set.
This issue can potentially be addressed by using a pre-solver.


\noindent \textbf{RQ4} \textit{Which algorithm selection method is the best?}
%
As reviewed in \cite{LindauerHHS15}, various algorithm selection methods have been proposed in the filed of artificial intelligence.
For example, the regression-based method \cite{XuHHL08} constructs a performance model for each algorithm and selects the best one from $\mathcal{A}$ in terms of the predicted performance.
However, it is not clear which algorithm selection method is the best for black-box optimization.
Although the study \cite{KerschkeT19} evaluated the performance of three selection methods, it did not show details of the comparison results.
Unlike \cite{JankovicPED21}, we are interested in the performance of algorithm selection methods rather than the performance of machine learning models.

\noindent \textbf{RQ5} \textit{How difficult is it to outperform the SBS?}
The studies \cite{BischlMTP12,AbellMT13,DerbelLVAT19,KerschkeT19,JankovicPED21,MunozK21} discussed the effectiveness of algorithm selection systems by comparing them with the SBS.
They also compared algorithm selection systems with the virtual best solver (VBS), which is an oracle that always selects the best optimizer from $\mathcal{A}$ on any given problem.
A comparison with SBS and VBS allows understanding ``how far'' an algorithm selection system is from them using, e.g., a relative deviation.
If an algorithm selection system $S$ outperforms the SBS in $\mathcal{A}$, the previous studies concluded that $S$ is effective.
However, the difficulty of outperforming the SBS has not been well understood.

\noindent \textbf{RQ6} \textit{How does the choice of algorithm portfolios influence the overall performance of algorithm selection systems?}
The study \cite{KerschkeT19} gave the following rule of thumb to construct a portfolio $\mathcal{A}$:   ``\textit{Ideally, the considered set should be as small and as complementary as possible and should include state-of-the art optimizers}''.
Since it is difficult to select the best one from too many candidates, the portfolio size should be as small as possible.
The performance of the VBS of $\mathcal{A}$ should also be as good as possible.
However, the influence of the VBS and the size of portfolios on the performance of algorithm selection systems is unclear for black-box optimization.
It is also unclear which algorithm portfolio should be used in practice.


\noindent \textbf{Outline}

Section \ref{sec:preliminaries} provides some preliminaries.
Section \ref{sec:review} reviews previous studies.
Section \ref{sec:proposed_method} explains our approaches for benchmarking algorithm selection systems.
Section \ref{sec:setting} describes our experimental setting.
Section \ref{sec:results} shows analysis results.
Section \ref{sec:conclusion} concludes this paper.

\noindent \textbf{Supplementary file}

This paper refers to a figure and a table in the supplementary file as Figure S.$*$ and Table S.$*$, respectively.

\noindent \textbf{Code availability}

The source code used in this study is available at \url{https://github.com/ryojitanabe/as_bbo}.

\section{Preliminaries}
\label{sec:preliminaries}

First, Section \ref{sec:bbob} describes the BBOB function set \cite{hansen2012fun} and the COCO data archive  (\url{https://numbbo.github.io/data-archive}).
Then, Section \ref{sec:ert} explains the following three performance measures for black-box optimization: the expected runtime (ERT) \cite{AugerH05a}, the relative ERT (relERT) \cite{BischlMTP12,KerschkeT19}, and the successful performance 1 (SP1) \cite{AugerH05a}.
Section \ref{sec:ert} also explains other performance measures.


\subsection{The BBOB function set and the COCO data archive}
\label{sec:bbob}

The (noiseless) BBOB function set \cite{hansen2012fun} consists of the 24 parameterized functions, which are grouped into the following five categories: separable functions ($f_1, ...,  f_5$),  functions with low or moderate conditioning ($f_6, ..., f_9$), functions with high conditioning and unimodal ($f_{10}, ..., f_{14}$), multimodal functions with adequate global structure ($f_{15}, ..., f_{19}$), and multimodal functions with weak global structure ($f_{20}, ...,  f_{24}$).
Each BBOB function represents one or more difficulties in real-world black-box optimization. 
Each BBOB function is instantiated with different parameters.


The BBOB workshop is held at the GECCO conference almost every year.
COCO \cite{HansenARMTB21} is a platform for benchmarking black-box optimizers.
The COCO data archive provides the benchmarking results of almost all optimizers that participated in the BBOB workshop.
Currently, except for incomplete results, the benchmarking results of 209 optimizers are available at the COCO data archive.
%
The number of instances for each BBOB function is fixed to 15 for all years.
However, as summarized in \cite{KerschkeT19}, only the first 5 out of 15 instances are commonly used in all years.
For this reason, most previous studies on algorithm selection used only the first five instances whose instance IDs are 1, 2, 3, 4, and 5.


\subsection{Performance measures for black-box optimization}
\label{sec:ert}


\subsubsection{ERT}

In the context of black-box numerical optimization, the runtime is generally measured in terms of the number of function evaluations rather than the computation time.
The ERT \cite{AugerH05a} measures the expected number of function evaluations needed to reach a target value $f_{\mathrm{target}} = f(\vector{x}^*) + \epsilon$, where $\vector{x}^*$ is the optimal solution, and $\epsilon$ is a precision level.
See Section \ref{sec:setting} for the $\epsilon$ value used in this study.
Note that most black-box optimizers (e.g., DE \cite{StornP97} and CMA-ES \cite{Hansen16a}) are invariant in terms of order-preserving transformations of the objective function value \cite{Hansen00}.


Suppose that an independent run of an optimizer $a$ is performed for each of instances of a function $f$.
In this case, the ERT value of $a$ is calculated as follows:
\begin{align}
  \label{eqn:ert}
  \mathrm{ERT} = \frac{\sum^{N^{\mathrm{run}}}_{i=1} \mathrm{FE}_i}{N^{\mathrm{succ}}},    
\end{align}
where $\mathrm{FE}_i$ is the number of all function evaluations conducted in the $i$-th function instance until $a$ terminates.
Here, $a$ immediately terminates when $a$ reaches $f_{\mathrm{target}}$.
$N^{\mathrm{run}}$ in \eqref{eqn:ert} is the number of runs, where $N^{\mathrm{run}}$ also represents the number of function instances in this study.
$N^{\mathrm{succ}}$ is the number of successful runs.
We say that a run of $a$ on the $i$-th function instance is successful if $a$ reaches $f_{\mathrm{target}}$.

\subsubsection{relERT}


The relERT \cite{BischlMTP12,KerschkeT19} is a normalized ERT using the ERT value of the best optimizer (bestERT) in $\mathcal{A}$ as follows: relERT$=$ERT$/$bestERT.
Here, the best optimizer is determined based on its ERT value for all instances of the corresponding function.
%
The ERT values significantly differ depending on the difficulty of a function.
The relERT aims to evaluate the performance of optimizers on the same scale.

The ERT value in \eqref{eqn:ert} and the relERT value are not computable when all runs of $a$ are unsuccessful (i.e., $N^{\mathrm{succ}}=0$).
The previous studies \cite{BischlMTP12,AbellMT13,DerbelLVAT19,KerschkeT19} imputed the missing relERT value using the penalized average runtime (PAR10) score \cite{LindauerRK19}.
Since PAR10 was used in many previous studies for algorithm selection (e.g., \cite{BischlMTP12,BischlKKLMFHHLT16,KerschkeKBHT18,KerschkeT19,DerbelLVAT19,LindauerRK19}), we adopted PAR10.
Similarly, we replaced the missing relERT value with ten times the worst relERT ($\mathrm{relERT}^{\mathrm{worst}}$) value of all optimizers in $\mathcal{A}$ for each $n$.
Precisely, for each dimension $n$, we defined the $\mathrm{relERT}^{\mathrm{worst}}$ value based on the relERT values of all algorithms in $\mathcal{A}$ on all the 24 BBOB functions ($f_1, \dots, f_{24}$) as follows: $\mathrm{relERT}^{\mathrm{worst}} = \max_{a \in \mathcal{A}, f \in \{f_1, ..., f_{24}\}} \{\mathrm{relERT}(a, f) \}$, where $\mathrm{relERT}(a, f)$ is the relERT value of $a$ on the $n$-dimensional $f$.


\subsubsection{SP1}

Similar to the ERT, the SP1 \cite{AugerH05a} estimates the expected number of function evaluations to reach $f_{\mathrm{target}}$.
The SP1 assumes that the expected number of function evaluations for unsuccessful runs equals that for successful runs \cite{AugerHZRS09}.
Unlike the ERT, the SP1 is not sensitive to the maximum number of function evaluations.
The SP1 is defined as follows:
\begin{align}
  \label{eqn:sp1}
  \mathrm{SP1} = \frac{\mathrm{FE}^{\mathrm{avg}}}{p^{\mathrm{succ}}},    
\end{align}
where $\mathrm{FE}^{\mathrm{avg}}$ is the average number of function evaluations for successful runs.
In \eqref{eqn:sp1}, $p^{\mathrm{succ}}$ is the success probability, which is the number of successful runs $N^{\mathrm{succ}}$ divided by the number of runs $N^{\mathrm{run}}$ (i.e., $p^{\mathrm{succ}} = N^{\mathrm{succ}} / N^{\mathrm{run}}$).

\subsubsection{Notes on ERT, relERT, and SP1}
\label{sec:notes_measures}

Here, we describe some notes on ERT, relERT, and SP1.
The three measures require $f_{\mathrm{target}}$, which can depend on $f(\vector{x}^*)$.
Since $f(\vector{x}^*)$ is unknown in most real-world problems, the three measures are not always available.
Each optimizer has one or more stopping conditions, e.g., the maximum budget of evaluations $b^{\mathrm{max}}$.
Since the ERT in \eqref{eqn:ert} takes into account the number of function evaluations used in an unsuccessful run, the ERT is sensitive to a stopping condition of an optimizer.
Section 3.3.2 in \cite{DerbelLVAT19} shows how sensitive the ERT is to $b^{\mathrm{max}}$ using an intuitive example.
The same is true for the relERT.
An optimizer can possibly reach $f_{\mathrm{target}}$ when setting $b^{\mathrm{max}}$ to a sufficiently large number, and vice versa.
In other words, the results of an optimizer depend on $b^{\mathrm{max}}$.
Thus, any performance measure (including the SP1) is influenced by $b^{\mathrm{max}}$ even when considering the same optimizer.







\subsubsection{Other performance measures}

The ERT, relERT, and SP1 measures are for the fixed-target scenario, which is representative in the BBOB community.
The performance for the fixed-budget scenario is generally evaluated by the quality of the best-so-far solution \cite{JankovicD20}.
The study \cite{Rijn0SB17} proposed a hybrid measure of the ERT and the error value.

Some previous studies (e.g., \cite{Lopez-IbanezS14,JesusLDP20}) proposed anytime performance measures.
Roughly speaking, this kind of measure commonly aims to evaluate the anytime performance of an optimizer by calculating the ``volume'' of its performance profile.
For example, the IOHprofiler platform \cite{DoerrYHWSB20} provides an anytime performance measure based on the area under the curve of the empirical cumulative distribution function.




\section{Literature review}
\label{sec:review}

This section reviews previous studies on algorithm selection for black-box optimization problems as well as other problems.
%
First, Section \ref{sec:features} describes features for algorithm selection only for black-box numerical optimization.
Then, Section \ref{sec:ap} introduces methods for constructing algorithm portfolios.
Section \ref{sec:selection_methods} describes selection methods.
Section \ref{sec:cv} explains cross-validation methods.
Section \ref{sec:presolver} describes pre-solvers.
Finally, Section \ref{sec:prev_studies} describes related work in other domains.

\subsection{Features}
\label{sec:features}


Feature-based algorithm selection systems require a set of domain-dependent features, which represent the characteristics of a given problem.
It is challenging to design helpful features for black-box optimization.
As discussed in \cite{AbellMT13}, unlike other problems (e.g., SAT), only scarce information about a problem is available from its definition.
For this reason, features need to be computed based on a set of $s$ solutions $\mathcal{X} = \{\vector{x}_i\}^s_{i=1}$ and their objective values $f(\mathcal{X})$, where the $s$ solutions can be generated by any method, e.g., random sampling and local search.
However, the evaluation of solutions by the objective function $f$ is generally computationally expensive.
Thus, features should be computed based on a small-size $\mathcal{X}$.

The ELA approach \cite{MersmannBTPWR11} is generally used to compute features for black-box optimization.
ELA computes a set of numerical features from $\mathcal{X}$ and $f(\mathcal{X})$.
%
Most previous studies used the \textsf{R}-package \texttt{flacco} \cite{KerschkeT2019flacco} to compute ELA features.
Table \ref{suptab:flacco_features} shows 17 feature classes currently provided by \texttt{flacco}.
Each feature class consists of more than one feature.
In total, 342 features are available in \texttt{flacco}.
However, three feature classes (\texttt{ela\_conv}, \texttt{ela\_curv}, and \texttt{ela\_local}) have not been used in most recent studies since they need additional function evaluations apart from $\mathcal{X}$.
Since the five cell mapping feature classes (e.g., \texttt{cm\_angle}) are computable only for $n \leq 5$, they have not been generally used.




Each feature class characterizes the properties of a problem in a different way.
For example, the \texttt{ela\_meta} feature class builds multiple regression models based on $\mathcal{X}$ and $f(\mathcal{X})$.
The model-fitting results are the \texttt{ela\_meta} features, e.g., how well the regression models can fit the given data set.
The \texttt{disp} features are computed by the dispersion metric \cite{LunacekW06}, which was designed for quantifying the degree of the global structure.

\subsection{Construction methods of algorithm portfolios}
\label{sec:ap}

Since no optimizer can perform the best on all function instances, it is necessary to select the most promising optimizer from a portfolio on a given instance.
We explain three methods \cite{BischlMTP12,MunozK16,KerschkeHNT19} for constructing a portfolio of $k$ algorithms $\mathcal{A} = \{a_1, ...,  a_k\}$ for algorithm selection.
We do not describe methods for construing \textit{parallel algorithm portfolio approaches} (e.g., \cite{TangPCY14}).
The three construction methods aim to construct a portfolio whose optimizer can solve at least one BBOB function.
Even when a portfolio includes ``a correct answer'', algorithm selection systems cannot know it due to the exclusive property of the cross-validation methods (see Section \ref{sec:cv}).
%
In real-world applications, an algorithm selection system selects the most promising algorithm from a portfolio on a given problem.
\textsc{Hydra} \cite{XuHL10,XuHHL11} is an efficient automatic configurator for algorithm portfolios for the combinatorial domain.
Although \textsc{Hydra} requires a parameterized system, how to construct it in the continuous domain is unclear.
An extension of \textsc{Hydra} to the continuous domain is also beyond the scope of this paper.
%
Section \ref{sec:results} investigates the influence of $k$ on the performance of algorithm selection systems later.

In \cite{BischlMTP12}, first, a set of candidates $\mathcal{R}$ are selected so that each algorithm in $\mathcal{R}$ performs the best on at least one BBOB function in terms of the ERT.
Then, $k$ algorithms in $\mathcal{A}$ are further selected from $\mathcal{R}$ so that $\mathcal{A}$ minimizes the worst-case performance of the VBS on the 24 BBOB functions.
Here, the study \cite{BischlMTP12} did not explain how to get $\mathcal{A}$ from $\mathcal{R}$.
%
A method proposed in \cite{MunozK16} first constructs $\mathcal{R}$ as in \cite{BischlMTP12}.
Then, the method iteratively selects an algorithm from $\mathcal{R}$ based on a voting strategy.
Here, the method uses the ERT as a performance measure.
A method proposed in \cite{KerschkeHNT19} constructs a portfolio on the BBOB functions with $n \in \{2, 3, 5, 10\}$.
First, all candidates are ranked based on their ERT values.
Then, four candidate sets $\mathcal{R}_2$, $\mathcal{R}_3$, $\mathcal{R}_5$, and $\mathcal{R}_{10}$ are constructed for 2, 3, 5, and 10 dimensions, respectively.
$\mathcal{R}_{n}$ contains algorithms ranked within the top 3 of at least one $n$-dimensional BBOB function.
Finally, the method selects algorithms that commonly belong to the four sets, i.e., $\mathcal{A} = \bigcap_{n \in \{2, 3, 5, 10\}} \mathcal{R}_n$.


\subsection{Algorithm selection methods}
\label{sec:selection_methods}

An algorithm selection method aims to find a mapping from features to $k$ algorithms in a portfolio $\mathcal{A} = \{a_1, ..., a_k\}$ by machine learning.
As reviewed in \cite{LindauerHHS15}, various algorithm selection methods have been proposed in the field of artificial intelligence.
In \cite{LindauerHHS15}, Lindauer et al. proposed \textsc{AutoFolio}, which is a highly-parameterized algorithm selection framework for combinatorial optimization problems, including SAT and ASP.
They generalized existing algorithm selection methods for \textsc{AutoFolio}.
Inspired by \cite{LindauerHHS15}, this paper considers the following five general algorithm selection methods:

\subsubsection{Classification}

The classification-based method was used in LLAMA \cite{Kotthoff13} and previous studies for black-box optimization \cite{BischlMTP12,KerschkeT19}.
The classification-based method builds a classification model to directly predict the best algorithm from the $k$ algorithms in $\mathcal{A}$ based on a given feature set, where the best algorithm is determined for each function based on a given performance measure.
As pointed out in \cite{KerschkeT19}, the classification-based method does not consider the performance rankings of the other $k-1$ algorithms.

\subsubsection{Regression}
\label{sec:regression_selector}

The regression-based method was used in \textsc{SATzilla}'09 \cite{XuHHL08} and recent studies for black-box optimization \cite{KerschkeT19,DerbelLVAT19,JankovicPED21}.
First, the regression-based method constructs $k$ regression models for $k$ algorithms in $\mathcal{A}$, respectively.
Then, the regression-based method selects the best one from $k$ algorithms based on their predicted performance.

\subsubsection{Pairwise classification}

The pairwise classification-based method was used in \textsc{SATzilla}'11 \cite{XuHHL11}. 
The study \cite{LindauerHHS15} reported the promising performance of the pairwise classification-based method for combinatorial optimization.
A similar selection method was also adopted by a study for black-box optimization \cite{MunozK21}.
In the training phase, the pairwise classification-based method builds a classification model for each pair of $k$ algorithms in $\mathcal{A}$.
In the testing phase, the method evaluates all $\binom{k}{2}$ models.
Then, the method selects the best one out of $k$ algorithms in terms of the number of votes.
In this study, ties are broken randomly.

\subsubsection{Pairwise regression}

Although the pairwise regression-based method was originally proposed for the TSP \cite{KerschkeKBHT18}, it has been used for black-box optimization \cite{KerschkeT19}.
First, the method constructs a regression model for each pair of $k$ algorithms in $\mathcal{A}$, where the model predicts the performance difference between two algorithms for each pair.
Then, the method selects the best one from $k$ algorithms based on the sum of predicted performance differences.

\subsubsection{Clustering}

The clustering-based method was used in ISAC \cite{KadiogluMST10}.
In the training phase, for each feature, feature values of function instances are normalized in the range $[-1, 1]$.
Then, the $g$-means \cite{HamerlyE03} clustering of function instances is performed based on their normalized feature values, where $g$-means automatically determines an appropriate number of clusters.
In the testing phase, an unseen function instance is assigned to the nearest cluster based on its normalized feature values.
Finally, the best algorithm in the nearest cluster is selected.
This study determines the best algorithm in each cluster according to the average ranking based on a performance measure.

\subsection{Cross-validation methods}
\label{sec:cv}


Algorithm selection will be ultimately performed on a real-world problem.
Thus, the performance of algorithm selection systems should be evaluated in an unbiased manner.
For this purpose, most previous studies (e.g., \cite{BischlMTP12,DerbelLVAT19,KerschkeT19,JankovicPED21,MunozK21}) used cross-validation methods for benchmarking algorithm selection systems for black-box optimization.

Below, we explain the following four cross-validation methods used in the literature: leave-one-instance-out cross-validation (LOIO-CV) \cite{BischlMTP12,JankovicPED21}, leave-one-problem-out cross-validation (LOPO-CV) \cite{BischlMTP12,DerbelLVAT19}, leave-one-problem-out-across-dimensions cross-validation (LOPOAD-CV) \cite{KerschkeT19}, and 10-fold randomized-instance cross-validation (RI-CV).

As explained in Section \ref{sec:bbob}, this study considers only the first five instances for each BBOB function, i.e., $|\mathcal{I}_{i}| = 5$ for $f_i$ ($i \in \{1, ..., 24\}$).
As in \cite{KerschkeT19}, this paper sets $n$ to $2, 3, 5$, and $10$.
Let $\mathcal{I}^{\mathrm{train}}$ be a set of function instances used in the training phase.
Let also $\mathcal{I}^{\mathrm{test}}$ be a set of function instances used in the testing phase.
Note that setting described here depends on previous benchmarking studies on the BBOB function set.
Note also that the LOIO-CV, the LOPO-CV, and the LOPOAD-CV cannot always be applied to any function set.
For example, the LOIO-CV is inapplicable when the number of instances for each function is only one.



\subsubsection{LOIO-CV}

A 5-fold cross-validation is performed on the 24 BBOB functions for each dimension $n$.
In the $i$-th fold, $\mathcal{I}^{\mathrm{test}}$ is the set of the 24 $i$-th instances $\mathcal{I}_{i}$, where $|\mathcal{I}^{\mathrm{test}}| = 24 \times 1 = 24$.
Thus, $\mathcal{I}^{\mathrm{train}}$ is $\mathcal{I}_{1} \cup \cdots \cup \mathcal{I}_{5} \setminus \mathcal{I}_i$, where $|\mathcal{I}^{\mathrm{train}}| = 24 \times 4 = 96$.
Since function instances used in the training and testing phases are relatively similar, the LOIO-CV is likely easier than the LOPO-CV explained later.

For the LOIO-CV, the calculation of a performance measure value needs special care to avoid ``data leakage''.
Let us consider the calculation of the ERT value on a function $f$ in the $i$-th fold, where $i \in \{1, ..., 5\}$.
Generally, test datasets should not be available in the training phase.
For this reason, the ERT value in the training phase should be calculated based only on the remaining four instances ($\{1, ..., 5\} \setminus i$).

\subsubsection{LOPO-CV}

A 24-fold cross-validation is performed on the 24 BBOB functions for each dimension $n$.
In the $i$-th fold, $\mathcal{I}^{\mathrm{test}}$ is the set of the five instances of the $i$-th function $f_i$, where $|\mathcal{I}^{\mathrm{test}}| = 1 \times 5 = 5$.
Thus, $\mathcal{I}^{\mathrm{train}}$ is $\mathcal{I}_{1} \cup \cdots \cup \mathcal{I}_{24} \setminus \mathcal{I}_i$, where $|\mathcal{I}^{\mathrm{train}}| = 23 \times 5 = 115$.
Since the 24 BBOB functions have different properties, instances used in the training and testing phases can be quite dissimilar for the LOPO-CV.
For this reason, it is expected that algorithm selection for the LOPO-CV is challenging.

\subsubsection{LOPOAD-CV}

While the LOPO-CV is performed for each $n$, the LOPOAD-CV is performed across all four dimensions ($n \in \{2, 3, 5, 10\}$).
In the LOPOAD-CV, a 96-fold cross-validation is conducted on the 24 BBOB functions with all 4 dimensions, where $24 \times 4 = 96$.
%
Let $\mathcal{I}_{j,l}$ be the set of the five instances of the $j$-th function $f_{j}$ with the $l$-th dimension.
%
In the $(j \times l)$-th fold, $\mathcal{I}^{\mathrm{test}}$ is $\mathcal{I}_{j,l}$, where $|\mathcal{I}^{\mathrm{test}}| = 1 \times 5 = 5$.
Thus, $\mathcal{I}^{\mathrm{train}}$ is $\mathcal{I}_{1,1} \cup \cdots \cup \mathcal{I}_{24,4} \setminus \mathcal{I}_{j,l}$, where $|\mathcal{I}^{\mathrm{train}}| = 95 \times 5 = 475$.

Unlike the LOPO-CV and the LOIO-CV, the LOPOAD-CV evaluates the performance of algorithm selection systems on multiple dimensions.
If both the performance of algorithms in $\mathcal{A}$ and features have good scalability with respect to dimension, the LOPOAD-CV may be easier than the LOPO-CV.
This is because both $\mathcal{I}^{\mathrm{test}}$ and $\mathcal{I}^{\mathrm{train}}$ include instances of the same function.



\subsubsection{RI-CV}

A 10-fold random cross-validation is performed on the 24 BBOB functions for each dimension $n$.
The $120$ function instances ($24 \times 5 = 120$) are randomly grouped into 10 subsets of size 12 as follows: $\mathcal{I}_1, ..., \mathcal{I}_{10}$.
In the $i$-th fold, $\mathcal{I}^{\mathrm{test}}$ is the $i$-th subset $\mathcal{I}_{i}$, where $|\mathcal{I}^{\mathrm{test}}| = 12 \times 1 = 12$.
Thus, $\mathcal{I}^{\mathrm{train}}$ is $\mathcal{I}_{1} \cup \cdots \cup \mathcal{I}_{12} \setminus \mathcal{I}_i$, where $|\mathcal{I}^{\mathrm{train}}| = 12 \times 9 = 108$.
A 10-fold random cross-validation method has been generally used for algorithm selection in the combinatorial domain \cite{BischlKKLMFHHLT16}.
In contrast, only the study \cite{AbellMT13} used the 10-fold random cross-validation method for black-box optimization, where it did not describe details of the cross-validation method.
In the best case, function instances used in the training and testing phases can be similar as in the LOIO-CV.




\subsection{Pre-solvers}
\label{sec:presolver}


The concept of pre-solving was first adopted in \textsc{SATzilla} \cite{XuHHL08}, which is a representative algorithm selection system for SAT.
Pre-solving was also incorporated into some modern algorithm selection systems for combinatorial optimization, e.g., \textsc{claspfolio 2} \cite{HoosLS14} and 3S \cite{KadiogluMSSS11}.

Before starting the algorithm selection process, \textsc{SATzilla} runs two pre-defined pre-solvers on a given SAT instance within a short amount of time.
Only when the pre-solvers cannot solve the instance, \textsc{SATzilla} performs algorithm selection.
%
If the pre-solvers can quickly solve easy instances, an algorithm selection system can focus only on hard instances.
For some easy instances, the computation time of the algorithm selection process (including feature computation) dominates the runtime of a selected algorithm.
The use of the pre-solvers can avoid such unnecessary computation time for algorithm selection on easy instances.

\subsection{Related work in other problem domains}
\label{sec:prev_studies}

Below, we explain difficulties in algorithm selection for black-box optimization.
A review of all previous studies for algorithm selection is beyond the scope of this paper.
Interested readers can refer to exhaustive survey papers \cite{Smith-Miles08,Kotthoff16,KerschkeHNT19}.


It is difficult to compare the performance of algorithm selection systems on across different problem domains (e.g., SAT and ASP) in a common platform.
A benchmark library for algorithm selection (ASlib) \cite{BischlKKLMFHHLT16} addresses this issue by providing scenarios for various problem domains, which are represented in a standardized format.
ASlib was used in the algorithm selection competitions in 2015 and 2017 \cite{LindauerRK19}.
%

%


In addition to the performance sensitivity of optimizers and features (see the beginning of Sections \ref{sec:introduction} and \ref{sec:features}, respectively), the similarity of problem instances in the training and testing phases may be the main difference between algorithm selection for black-box numerical optimization and that for combinatorial optimization.
In most scenarios (e.g., for SAT and the TSP), at least some problem instances in the training and testing phases were taken from the same distribution of problem instances even when using heterogeneous instance distribution.
In contrast, as in the LOPO-CV, function instances used in the training and testing phases can be quite dissimilar for algorithm selection for black-box numerical optimization.
We discuss the rational reason to adopt the LOPO-CV in Section \ref{sec:exp_cv} later.

\section{Our approaches}
\label{sec:proposed_method}


First, Section \ref{sec:discussion_sfe} discussed the importance of considering the number of function evaluations in the sampling phase.
Then, Section \ref{sec:per_measure} explains that the relERT can overestimate the performance of algorithm selection systems.
Section \ref{sec:ls_ap} presents a local search method for constructing an algorithm portfolio of any size $k$.
Finally, Section \ref{sec:presolver_bbo} describes pre-solvers for black-box optimization.

\subsection{Necessity of considering the number of function evaluations in the sampling phase}
\label{sec:discussion_sfe}

As explained in Section \ref{sec:features}, the sample of $s$ solutions $\mathcal{X} = \{\vector{x}_i\}^s_{i=1}$ and their objective values $f(\mathcal{X})$ are needed to compute landscape features.
Thus, the sampling phase requires $s$ function evaluations. 
As shown in Table \ref{suptab:prev_as_systems}, the previous studies \cite{BischlMTP12,DerbelLVAT19,KerschkeT19,JankovicPED21,MunozK21} set $s$ to different numbers.
For example, the two previous studies \cite{BischlMTP12} and \cite{KerschkeT19} set $s$ to $500 \times n$ and $50 \times n$, respectively.
The start of the run of a selected optimizer should be delayed by $s$ function evaluations used in the sampling phase.
When $s$ is too large, the overall performance of an algorithm selection system can deteriorate.
For this reason, $s$ is an important factor for benchmarking of algorithm selection systems.
However, except for \cite{KerschkeT19}, most previous studies did not include the number of $s$ evaluations in the total number of function evaluations.
Note that most previous studies for algorithm selection in the combinatorial domain (e.g., \cite{XuHHL08,HoosLS14}) included the computation time of features in the total computation time of algorithm selection.

On the one hand, as discussed in \cite{KerschkeT19}, the sample $\mathcal{X}$ can be reused for the initial population of a population-based evolutionary algorithm (e.g., DE \cite{StornP97}).
The number of $s$ function evaluations may be negligibly small when a selected optimizer requires a much large number of function evaluations, e.g., on hard instances.
%
On the other hand, $\mathcal{X}$ \textit{cannot} be reused when an algorithm selection system selects a non-evolutionary algorithm or a model-based evolutionary algorithm (e.g., CMA-ES \cite{Hansen16a}).
The number of $s$ function evaluations is \textit{not negligible} when a selected optimizer requires a small number of function evaluations, e.g., on easy instances.
In any case, there is no rational reason to ignore $s$ function evaluations in the sampling phase.
Based on the above discussions, as in \cite{KerschkeT19}, this study includes the number of $s$ function evaluations in the total number of function evaluations.
In this case, the setting of $s$ influences the overall performance of algorithm selection systems.


\subsection{Undesirable property of the relERT}
\label{sec:per_measure}


Except for \cite{JankovicPED21}, all the previous studies used the relERT for evaluating the performance of algorithm selection systems in the fixed-target scenario.
It has been believed that the lower bound of the relERT value is 1.
However, we point out that the true lower bound of the relERT value is $1/$bestERT.
There are three cases where the relERT value is less than 1.

First, it has been assumed that $f(\vector{x}) > f_{\mathrm{target}}$ for any $\vector{x} $ in the sample $\mathcal{X}$.
In addition, none of the previous studies considered a pre-solver. 
When the sample $\mathcal{X}$ or a pre-solver reaches $f_{\mathrm{target}}$ faster than the best optimizer in a portfolio $\mathcal{A}$, an algorithm selection system obtains a lower ERT value than the bestERT value.
In this case, it is possible to achieve a relERT value of less than 1.

Second, it has been assumed that the same optimizer is selected from $\mathcal{A}$ for all the five instances of a function $f$.
In contrast, it is possible that different optimizers are selected from $\mathcal{A}$ for the five instances, respectively.
Note that the best optimizer mentioned here is determined based on the number of function evaluations to reach $f_{\mathrm{target}}$ on each instance.
When an algorithm selection system selects the best optimizer for each function instance in terms of the number of function evaluations, it obtains a relERT value of less than 1.

Third, apart from the above two ``nice mistakes'', a relERT value can accidentally be less than 1.
Let us consider a portfolio $\mathcal{A}$ of two algorithms $a_1$ and $a_2$.
While the maximum number of function evaluations in $a_1$ is $50$, that in $a_2$ is $5$.
Suppose that $a_1$ reached $f_{\mathrm{target}}$ within $30$ and $20$ on the first two out of the five instances of $f$ respectively, and $a_1$ failed to reach $f_{\mathrm{target}}$ on the other three instances.
Suppose also that all five runs of $a_2$ were unsuccessful.
In this case, the ERT value of $a_1$ is $(30+20+50 \times 3) / 2 = 100$, and that of $a_2$ is not computable.
Note that the missing relERT value of $a_2$ is imputed by the PAR10 score described in Section \ref{sec:ert}.
Thus, the bestERT value is $100$.
If an algorithm selection system selects $a_1$ on the first instance and $a_2$ on the other four instances, the ERT value of the system is $(30+5 \times 4) / 1 = 50$.
As a result, the system achieves the relERT value of $0.5 $ $(=50/100)$, even though it actually failed to select the best algorithm.
One may incorrectly conclude that the system is two times faster than $a_1$ in solving $f$.
Section \ref{sec:exp_ert} shows a practical example later.

The third undesirable case is due to the sensitivity of the ERT to the maximum number of function evaluations for an unsuccessful run.
When all algorithms in $\mathcal{A}$ use the same maximum number of function evaluations, the third case never occurs.
However, it is not realistic to assume such a termination condition.

Based on the above discussion, this study uses the SP1, instead of the ERT.
As explained in Section \ref{sec:ert}, the SP1 is robust to the maximum number of function evaluations.
We do not claim that the SP1 is a better measure than the ERT \textit{for any purpose}.
Instead, we claim that the SP1 is more appropriate than the ERT \textit{for benchmarking algorithm selection systems}.
Inspired by the relERT, this study uses the relative SP1 (relSP1), which is the SP1 value of an algorithm in $\mathcal{A}$ divided by the SP1 value of the best algorithm in $\mathcal{A}$ on a function.
This is the same as the procedure for obtaining the best ERT described in Section \ref{sec:ert}.
This study also applies the PAR10 to a missing relSP1 value.

\subsection{Local search for constructing algorithm portfolios}
\label{sec:ls_ap}

To analyze the influence of the portfolio size on the performance of algorithm selection systems, this study requires algorithm portfolios of any size $k$.
Although the three construction methods reviewed in Section \ref{sec:ap} are available, they are not appropriate for this purpose.
The two methods proposed in \cite{MunozK16,KerschkeT19} cannot control the size of $\mathcal{A}$.
Unlike the method proposed in \cite{BischlMTP12}, our method aims to optimize the VBS performance of $\mathcal{A}$.
Note that our method is only for an analysis of algorithm portfolios.
Thus, we do not claim that our method is more effective than the three existing methods.

Let $\mathcal{R}$ be a set of $l$ optimizers, where $l=209$ in this study.
Let also $\mathcal{A}$ be a portfolio of $k$ optimizers, where $k < l$.
As pointed out in \cite{MunozK16}, constructing $\mathcal{A}$ can be defined as a subset selection problem.
The goal of the problem is to select $\mathcal{A} \subset \mathcal{R}$ that minimizes a quality measure $m: \mathcal{A} \times  \mathcal{I} \rightarrow \mathbb{R}$ on a set of problem instances $\mathcal{I}$.
Fortunately, a general local search method for the subset selection problem can be applied to this portfolio construction problem in a straightforward manner.
This study uses a first-improvement local search method \cite{BasseurDGL16}, which was proposed for the hypervolume subset selection problem.
Algorithm \ref{supalg:ls} shows the local search method.
First, the local search method initializes $\mathcal{A}$ by randomly selecting $k$ optimizers from $\mathcal{R}$.
Then, for each iteration, the local search method swaps a pair of optimizers $a \in \mathcal{A}$ and $a' \in \mathcal{R} \setminus \mathcal{A}$.
The swap operation is performed until there is no pair to improve the following ranking-based quality measure $m$:
%
\begin{align}
  \label{eqn:low-risk}
  m(\mathcal{A}) &= \sum_{n \in \{2, 3, 5, 10\}} \sum_{i \in \{1, ..., 24\}} \mathrm{score}(\mathcal{A}, f^n_i),\\
  \label{eqn:score}
  \mathrm{score}(\mathcal{A}, f) &= \begin{cases}
   c \times  \underset{a \in \mathcal{A}}{\min} \{\mathrm{rank}(a, f)\} & \:  {\rm if} \exists a \: {\rm solves} \: f \\
    1 & \:  {\rm otherwise}
 \end{cases},
\end{align}
where $f^n_i$ in equation \eqref{eqn:low-risk} is the $n$-dimensional $i$-th BBOB function.
We rank $l$ optimizers in $\mathcal{R}$ based on their SP1 values for each $f^n_i$.
In equation \eqref{eqn:score}, $\mathrm{rank}(a, f)$ is a ranking of $a$ on $f$ in $\mathcal{R}$.
In equation \eqref{eqn:score}, $c$ is a coefficient value.
We set $c$ to $1/(l \times 24 \times 4)$ for the 24 BBOB functions with $n \in \{2, 3, 5, 10\}$.
If no $a$ in $\mathcal{A}$ reaches the target value, a penalty value of $1$ is assigned to $\mathrm{score}(\mathcal{A}, f)$.
If $m(\mathcal{A}) \geq 1$ in equation \eqref{eqn:low-risk}, it means that no optimizer in $\mathcal{A}$ could reach the target value $f_{\mathrm{target}}$ for at least one function.





\subsection{Pre-solvers for black-box optimization}
\label{sec:presolver_bbo}

It is straightforward to incorporate the concept of pre-solving into an algorithm selection system for black-box optimization.
In the pre-solving phase, a pre-solver is applied to a given problem with a small budget of function evaluations (e.g., $50 \times n$ evaluations) before the algorithm selection process starts.
Note that it is possible to use more than one pre-solvers as in \textsc{SATzilla} \cite{XuHHL08}.
When the pre-solver reaches the target value $f_{\mathrm{target}}$ on the problem, algorithm selection is not performed.
In this case, the algorithm selection system does not need to generate the sample $\mathcal{X}=\{\vector{x}_i\}^s_{i=1}$ of size $s$ to compute features.
In other words, $s$ function evaluations can be saved.
We expect that using a pre-solver can improve the performance of algorithm selection systems on easy instances.

It is desirable that a pre-solver can reach $f_{\mathrm{target}}$ on an easy instance with a small number of function evaluations.
This study uses SLSQP \cite{Kraft88} and SMAC \cite{HutterHL11} as pre-solvers.
Since evolutionary algorithms (e.g., DE and CMA-ES) generally require a relatively large number of function evaluations to find a good solution, they are not appropriate as pre-solvers.
In \cite{Hansen19}, Hansen showed the excellent performance of SLSQP on the BBOB functions for a small number of function evaluations.
SMAC is also a representative Bayesian optimization approach, which performs particularly well for computationally expensive optimization.
Thus, SLSQP and SMAC are reasonable first choices.


\section{Experimental setup}
\label{sec:setting}

This section explains the experimental setup.
Unlike previous studies, we performed 31 independent runs of algorithm selection systems, including pre-solving, sampling, feature computation, and cross-validation.
We conducted all experiments on a workstation with an Intel(R) 52-Core Xeon Platinum 8270 (26-Core$\times 2$) 2.7GHz and 768GB RAM using Ubuntu 18.04.
As in \cite{KerschkeT19}, we used the 24 noiseless BBOB functions \cite{hansen2012fun} with dimensions $n \in \{2, 3, 5, 10\}$.
We conducted our experiment by using the COCO platform \cite{HansenARMTB21}.
We also used the benchmarking results of 209 optimizers provided by the COCO data archive.
As in \cite{KerschkeT19,DerbelLVAT19}, we set a precision level $\epsilon$ to $10^{-2}$ for the ERT and SP1 calculations.
Note that the same $\epsilon$ value is used for pre-solvers.
The study \cite{MunozKH12} showed that the setting of $\epsilon$ does not significantly influence the accuracy of performance predictors.

%
We used the following nine non-expensive and scalable \texttt{flacco} feature classes in Table \ref{suptab:flacco_features}: \texttt{ela\_distr}, \texttt{ela\_level}, \texttt{ela\_meta}, \texttt{nbc}, \texttt{disp}, \texttt{ic}, \texttt{basic}, \texttt{limo}, and \texttt{pca}.
We employed \texttt{pflacco}, which is the Python interface of \texttt{flacco} (\url{https://github.com/Reiyan/pflacco}).
Our preliminary results showed that feature selection deteriorates the performance of algorithm selection systems in some cases.
This is consistent with the results reported in \cite{JankovicED21}.
Except for the classification-based selection method, wrapper feature selection approaches require extremely high computational cost as the portfolio size increases.
For these reasons, as in \cite{BischlMTP12,AbellMT13,DerbelLVAT19,JankovicPED21,MunozK21}, we did not perform feature selection.


As in most previous studies, we used the improved Latin hypercube sampling \cite{BeachkofskiG02} to generate the sample $\mathcal{X} = \{\vector{x}_i\}^s_{i=1}$ for feature computation.
We employed the \texttt{lhs} function in the \texttt{flacco} package.
As in \cite{KerschkeT19}, we set $s$ to $50 \times n$ unless explicitly noted, where the study \cite{KerschkeT19} selected this $s$ value based on the results in \cite{KerschkePWT15}.
For pre-solvers, we employed the SciPy implementation of SLSQP and the SMAC3 implementation \cite{LindauerEFBDBRS22} of SMAC.
We used the default parameter settings, including the termination criteria for SLSQP.
The maximum number of function evaluations was set to $50 \times n$, which is the same as the sampling phase.

We used random forest \cite{Breiman01} for the four supervised learning-based algorithm selection methods explained in Section \ref{sec:selection_methods}.
The random forest is a representative ensemble machine learning method that uses multiple decision trees.
Each tree is fitted to a subset of randomly selected features.
Then, the prediction is performed based on the number of votes by the trees.
Typical hyperparameters in the random forest include the number of trees (\texttt{n\_estimators}) and the feature subset size (\texttt{max\_features}).
As in ISAC \cite{KadiogluMST10}, we used $g$-means \cite{HamerlyE03} for the clustering-based selection method.
We employed the scikit-learn implementation of random forest and the pyclustering implementation of $g$-means with the default parameters, e.g., \texttt{n\_estimators}$=100$ and \texttt{max\_features}$=$\texttt{auto}.

Table \ref{tab:ap_ls} shows 14 algorithm portfolios used in Section \ref{sec:results}.
The first five portfolios ($\mathcal{A}_{\mathrm{kt}}$, ..., $\mathcal{A}_{\mathrm{mk}}$) were used in the previous studies \cite{KerschkeT19,DerbelLVAT19,JankovicPED21,BischlMTP12,MunozK21}, respectively.
We named each portfolio by taking the initial letters of the authors of the corresponding paper.
For each of the nine portfolios ($\mathcal{A}_{\mathrm{ls}2}$, ..., $\mathcal{A}_{\mathrm{ls}18}$), we performed 31 runs of the local search method explained in Section \ref{sec:ls_ap}.
Then, the best portfolio is selected in terms of $m$ in equation \eqref{eqn:low-risk}.
We confirmed that all the nine portfolios achieve $m$ values less than 1.

Sections \ref{sec:comparison_asmethods} and \ref{sec:exp_ap} use the performance score \cite{BaderZ11} to rank multiple algorithm selection systems.
For each dimension $n$, let us consider a comparison of $l$ algorithm selection systems $S_1, ..., S_l$ based on their 31 mean relSP1 values from 31 independent runs.
For $i \in \{1, ..., l\}$ and $ j \in \{1, ..., l\} \setminus \{i\}$, if $S_j$ performs significantly better than $S_i$ using the Wilcoxon rank-sum test with $p < 0.05$, then $\delta_{i,j} = 1$; otherwise, $\delta_{i,j} = 0$.
The score $P(S_i)$ is defined as follows: $P(S_i) = \sum_{ j \in \{1, ..., l\} \backslash \{i\}} \delta_{i,j}$.
The score $P(S_i)$ is the number of systems that outperform $S_i$ for each $n$.
A small $P(S_i)$ means that $S_i$ has a better performance among the $l$ systems.


In addition, we calculated the average rankings of algorithm selection systems by the Friedman test \cite{DerracGMH11}.
We used the CONTROLTEST package (\url{https://sci2s.ugr.es/sicidm}) to calculate the rankings.
However, the rankings by the Friedman test were generally consistent with those by the performance score.
For this reason, this paper shows only the latter results.

\definecolor{c1}{RGB}{150,150,150}
\definecolor{c2}{RGB}{220,220,220}

\section{Results}
\label{sec:results}

This section analyzes algorithm selection systems to answer the six research questions \textbf{RQ1}--\textbf{RQ6} discussed in Section \ref{sec:introduction}.
First, Section \ref{sec:exp_ert} shows a comparison of the ERT and the SP1 (\textbf{RQ1}).
Then, Section \ref{sec:exp_randomness} examines the influence of randomness on results of algorithm selection, including pre-solving, sampling, feature computation, and cross-validation  (\textbf{RQ2}).
Section \ref{sec:exp_presolver} investigates the effectiveness of SLSQP and SMAC as a pre-solver (\textbf{RQ3}).
Based on results shown in Section \ref{sec:exp_presolver}, Sections \ref{sec:comparison_asmethods}, \ref{sec:exp_cv}, and \ref{sec:exp_ap} use SLSQP as a pre-solver.
Section \ref{sec:comparison_asmethods} shows a comparison of the five algorithm selection methods  (\textbf{RQ4}).
Section \ref{sec:exp_cv} analyzes the difficulty of outperforming the SBS (\textbf{RQ5}).
Finally, Section \ref{sec:exp_ap} examines the performance of algorithm selection systems with the 14 algorithm portfolios shown in Table \ref{tab:ap_ls} (\textbf{RQ6}).

The influence of cross-validation methods on the results of algorithm selection systems is unclear.
For this reason, we here investigate it by using the four cross-validation methods. 
We essentially aim to analyze the influence of the similarity of function instances used in the training and testing phases.
However, it would be misleading to make conclusions based on the results achieved by multiple cross-validation methods.
Thus, we answer each question based only on the representative results for the LOPO-CV, which is the most practical (see the discussion in Section \ref{sec:exp_cv}).

\subsection{Comparison of the ERT and the SP1 (RQ1)}
\label{sec:exp_ert}

\begin{table}[t]
\setlength{\tabcolsep}{5.5pt} 
\centering
  \caption{\small Comparison of HCMA, HMLSL, and the regression-based algorithm selection system  (AS) using $\mathcal{A}_{\mathrm{kt}}$ on the 5-dimensional $f_{24}$ for the LOPO-CV. If the corresponding run was successful, the number in parentheses is 1. Otherwise, it is 0.
    }
\label{tab:ert_sp1}
{\scriptsize
  \begin{tabular}{lcccccccc}
  \toprule
 & HCMA & HMLSL & AS \#1 & AS \#2\\
  \midrule
FEs on $i_1$ & $3\,097\,698$ (1) & $100\,021$ (0) & $3\,097\,948$ (1) & $3\,097\,948$ (1)\\
FEs on $i_2$ & $2\,254\,446$ (1) & $100\,002$ (0) & $2254696$ (1) & $100\,252$ (0)\\
FEs on $i_3$ & $5\,001\,954$ (0) & $100\,206$ (0) & $100\,456$ (0) & $100\,456$ (0)\\
FEs on $i_4$ & $5\,001\,015$ (0) & $100\,008$ (0) & $100\,258$ (0) & $100\,258$ (0)\\
FEs on $i_5$ & $5\,001\,872$ (0) & $100\,006$ (0) & $100\,256$ (0) & $100\,256$ (0)\\
  \midrule
ERT & $10\,178\,492.50$ & Na & $2\,826\,807$ & $3\,499\,170$\\
SP1 & $6\,690\,180$ & Na & $6\,690\,805$ & $15\,489\,740$\\
  \midrule
relERT & $1$ & Na & $0.28$ & $0.34$\\
relSP1 & $1$ & Na & $1.00009$ & $2.32$\\
\bottomrule
  \end{tabular}
}
\end{table}

Table \ref{tab:ert_sp1} shows a comparison of HCMA, HMLSL, and the regression-based algorithm selection system using $\mathcal{A}_{\mathrm{kt}}$ on the 5-dimensional $f_{24}$ for the LOPO-CV.
Note that only this section uses the relERT, instead of the relSP1.
Table \ref{tab:ert_sp1} reports the number of function evaluations (FEs) used in the search on each of the five instances.
Table \ref{tab:ert_sp1} also reports the ERT, SP1, relERT, and relSP1 values.
HCMA performs the best on $f_{24}$ with $n=5$.
All five runs of HMLSL are unsuccessful on $f_{24}$ with $n=5$.
%
Table \ref{tab:ert_sp1} shows results of the best run (AS \#1) and the second best run  (AS \#2) out of the 31 runs of the system in terms the ERT (not the SP1).
For AS \#1 and AS \#2, the number of function evaluations includes $50 \times 5 = 250$ function evaluations in the sampling phase.

As shown in Table \ref{tab:ert_sp1}, AS \#1 selected HCMA on the first two instances and HMLSL on the other instances.
In contrast, AS \#2 selected HCMA on the first instance and HMLSL on the other instances.
Thus, both AS \#1 and AS \#2 failed to select HCMA on all five instances.
Nevertheless, AS \#1 and AS \#2 perform significantly better than HCMA in terms of the ERT.
AS \#1 and AS \#2 also achieve relERT values of less than 1.
%
As seen from Table \ref{tab:ert_sp1}, the maximum number of function evaluations of HCMA is approximately 50 times larger than that of HMLSL.
For this reason, when HMLSL is selected for unsuccessful runs, a better ERT value can be obtained.
This is exactly the third case explained in Section \ref{sec:per_measure}.

We observed the undesirable ERT (and relERT) results due to the third case only on $f_{24}$, where no optimizer in $\mathcal{A}_{\mathrm{kt}}$ can solve all the five instances of $f_{24}$.
However, as demonstrated here, the ERT and the relERT can possibly overestimate the performance of algorithm selection systems.
In contrast, as shown in Table \ref{tab:ert_sp1}, the SP1 and the relSP1 do not overestimate the performance of AS \#1  and AS \#2.
AS \#1 performs quite slightly worse than HCMA in terms of the SP1 and the relSP1.
The relSP1 of AS \#2 is 2.32 due to the unsuccessful run on the second instance.

\noindent \colorbox{c2}{\textbf{Answers to RQ1}}
We suggest using the SP1 and the relSP1 \textit{for benchmarking algorithm selection systems}, instead of the ERT and the relERT.
It should be highly noted that the above-discussed issue of the ERT never occurs for benchmarking a single optimizer and multiple optimizers with exactly the same termination conditions.






  %





\subsection{Influence of randomness (RQ2)}
\label{sec:exp_randomness}

\begin{figure}[t]
   \centering
\includegraphics[width=0.3\textwidth]{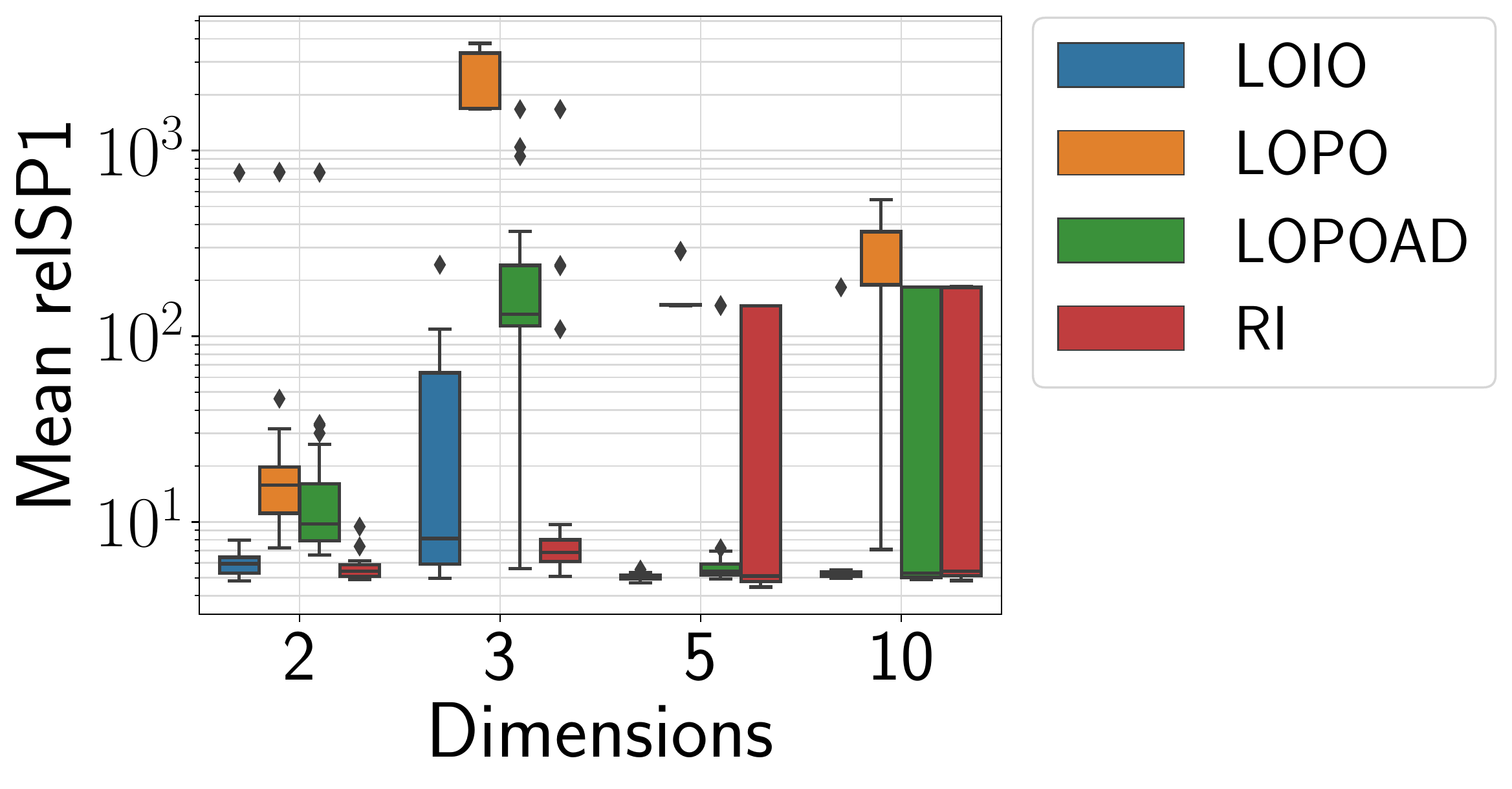}
\caption{Distributions of 31 mean relSP1 values of the pairwise classification-based algorithm selection system with $\mathcal{A}_{\mathrm{kt}}$.}
   \label{fig:dist_mean_relsp1}
\end{figure}

The previous studies \cite{BischlMTP12,AbellMT13,DerbelLVAT19,KerschkeT19} discussed the performance of algorithm selection systems based on the mean of the performance measure values (e.g., the relERT value) on the 24 BBOB functions for a single run.
In contrast, Fig. \ref{fig:dist_mean_relsp1} shows the distributions of 31 ``mean relSP1'' values, which were obtained by 31 runs of the pairwise classification-based system with $\mathcal{A}_{\mathrm{kt}}$ for the four cross-validation methods.
%
Figs. \ref{supfig:dist_meanrelsp1_kt}--\ref{supfig:dist_meanrelsp1_mk} show results of the five systems with the five portfolios, respectively.
We do not explain the results in Figs. \ref{supfig:dist_meanrelsp1_kt}--\ref{supfig:dist_meanrelsp1_mk} here, but they are almost consistent with Fig. \ref{fig:dist_mean_relsp1}.
Figs. \ref{supfig:dist_meanrelsp1_kt_nonlog}--\ref{supfig:dist_meanrelsp1_mk_nonlog} also show the non-log scale versions of Figs. \ref{supfig:dist_meanrelsp1_kt}--\ref{supfig:dist_meanrelsp1_mk}, respectively.


As shown in Fig. \ref{fig:dist_mean_relsp1}, the distribution of 31 ``mean relSP1'' values depends on the cross-validation method.
While the dispersion in the distribution of the ``mean relSP1'' values is relatively small for the LOIO-CV (except for the results on $n=3$), that is relatively large for the LOPO-CV, the LOPOAD-CV, and the RI-CV (except for the results on $n=5$).
For example, the minimum ``mean relSP1'' value on $n=10$ for the LOPO-CV is $7.10$, but the maximum one is $542.62$.
Thus, the best-case performance of the algorithm selection system is approximately $76$ times better than the worst-case one.
Figs. \ref{supfig:pc_as_loio}--\ref{supfig:pc_as_ri} show the distribution of ``relSP1'' (not ``mean relSP1'') values on the 24 BBOB functions for the four cross-validation methods, respectively.
Figs. \ref{supfig:pc_as_loio}--\ref{supfig:pc_as_ri} indicate that large variations in relSP1 values are due to unsuccessful selection on multimodal functions. 


Note that we cannot generalize our observation in Fig. \ref{fig:dist_mean_relsp1} for any portfolio.
For example, as seen from Fig. \ref{supfig:dist_meanrelsp1_bmtp}(c), when using $\mathcal{A}_{\mathrm{bmtp}}$, the dispersion in the distribution of the ``mean relSP1'' values is relatively large for the LOIO-CV.
Thus, our observations in Figs. \ref{fig:dist_mean_relsp1} and \ref{supfig:dist_meanrelsp1_bmtp}(c) are inconsistent with each other.
However, we can say that the influence of randomness is not negligible in any case.

\noindent \colorbox{c2}{\textbf{Answers to RQ2}}
We demonstrated that randomness influences results of algorithm selection systems in most cases.
This observation is consistent with the results for combinatorial optimization investigated in \cite{LindauerRK19}.
Because the best-case and worst-case results can be significantly different, we suggest performing multiple runs of algorithm selection systems.

\begin{table*}[t]
\setlength{\tabcolsep}{2.4pt} 
\centering
  \caption{\small Results of the pairwise classification-based algorithm selection system with and without the pre-solvers.
    }
\label{tab:effect_presolver}
{\scriptsize
\subfloat[LOIO-CV ($\mathcal{A}_{\mathrm{kt}}$)]{
\begin{tabular}{lllllll}
\toprule
System & $n=2$ & $n=3$ & $n=5$ & $n=10$\\
\midrule
AS ($50 \times n$) & 5.94 & \cellcolor{c1}8.13 & 5.01 & \cellcolor{c1}5.28\\
AS ($100 \times n$) & 8.75$-$ & 11.33$-$ & 7.88$-$ & 8.59$-$\\
SLSQP-AS & \cellcolor{c1}3.53$+$ & \cellcolor{c1}4.85$+$ & \cellcolor{c1}2.54$+$ & \cellcolor{c1}2.49$+$\\
SMAC-AS & 6.18 & \cellcolor{c1}7.82 & 7.07$-$ & 8.60$-$\\
\midrule
SBS & 5.81 & 9.82 & 4.49 & 6.76\\
\bottomrule
\end{tabular}
}
\subfloat[LOPO-CV ($\mathcal{A}_{\mathrm{kt}}$)]{
\begin{tabular}{lllllll}
\toprule
System & $n=2$ & $n=3$ & $n=5$ & $n=10$\\
\midrule
AS ($50 \times n$) & 15.75 & 3337.17 & 146.83 & 363.62\\
AS ($100 \times n$) & 16.66 & 3340.90 & 149.85$-$ & 366.78\\
SLSQP-AS & 11.58$+$ & 3334.35 & 144.06$+$ & 360.35\\
SMAC-AS & 15.59 & 3338.37 & 148.80$-$ & 366.99$-$\\
\midrule
SBS & 5.81 & 9.82 & 4.49 & 6.76\\
\bottomrule
\end{tabular}
}
\subfloat[LOPOAD-CV ($\mathcal{A}_{\mathrm{kt}}$)]{
\begin{tabular}{lllllll}
\toprule
System & $n=2$ & $n=3$ & $n=5$ & $n=10$\\
\midrule
AS ($50 \times n$) & 9.76 & 131.42 & 5.42 & \cellcolor{c1}5.29\\
AS ($100 \times n$) & 11.22$-$ & 124.18 & 8.01$-$ & 8.60$-$\\
SLSQP-AS & 7.36$+$ & 128.10 & \cellcolor{c1}2.92$+$ & \cellcolor{c1}2.50$+$\\
SMAC-AS & 9.68 & 115.13 & 7.54$-$ & 8.64$-$\\
\midrule
SBS & 5.81 & 9.82 & 4.49 & 6.76\\
\bottomrule
\end{tabular}
}
\\
\subfloat[LOIO-CV ($\mathcal{A}_{\mathrm{bmtp}}$)]{
\begin{tabular}{lllllll}
\toprule
System & $n=2$ & $n=3$ & $n=5$ & $n=10$\\
\midrule
AS ($50 \times n$) & \cellcolor{c1}7.41 & 36.15 & \cellcolor{c1}96.86 & \cellcolor{c1}158.07\\
AS ($100 \times n$) & \cellcolor{c1}7.82 & \cellcolor{c1}18.40$+$ & \cellcolor{c1}7.07 & \cellcolor{c1}160.06$-$\\
SLSQP-AS & \cellcolor{c1}5.55 & \cellcolor{c1}34.76 & \cellcolor{c1}95.28$+$ & \cellcolor{c1}156.52$+$\\
SMAC-AS & \cellcolor{c1}6.77 & 36.83 & \cellcolor{c1}97.93$-$ & \cellcolor{c1}159.64$-$\\
\midrule
SBS & 11.53 & 35.92 & 102.10 & 315.32\\
\bottomrule
\end{tabular}
}
\subfloat[LOPO-CV ($\mathcal{A}_{\mathrm{bmtp}}$)]{
\begin{tabular}{lllllll}
\toprule
System & $n=2$ & $n=3$ & $n=5$ & $n=10$\\
\midrule
AS ($50 \times n$) & 12.30 & 947.32 & 109.69 & 323.98\\
AS ($100 \times n$) & 14.23$-$ & 944.64 & 106.29 & 324.18\\
SLSQP-AS & \cellcolor{c1}10.20$+$ & 743.19 & 103.99 & \cellcolor{c1}314.43$+$\\
SMAC-AS & \cellcolor{c1}11.51 & 716.73 & 111.16 & 325.62\\
\midrule
SBS & 11.53 & 35.92 & 102.10 & 315.32\\
\bottomrule
\end{tabular}
}
\subfloat[LOPOAD-CV ($\mathcal{A}_{\mathrm{bmtp}}$)]{
\begin{tabular}{lllllll}
\toprule
System & $n=2$ & $n=3$ & $n=5$ & $n=10$\\
\midrule
AS ($50 \times n$) & 12.93 & \cellcolor{c1}13.28 & \cellcolor{c1}7.19 & \cellcolor{c1}158.48\\
AS ($100 \times n$) & 14.82$-$ & \cellcolor{c1}12.54 & \cellcolor{c1}7.41 & \cellcolor{c1}160.06$-$\\
SLSQP-AS & \cellcolor{c1}11.41$+$ & \cellcolor{c1}11.73 & \cellcolor{c1}5.29$+$ & \cellcolor{c1}156.84$+$\\
SMAC-AS & \cellcolor{c1}10.76$+$ & \cellcolor{c1}13.79 & \cellcolor{c1}8.27 & \cellcolor{c1}160.10$-$\\
\midrule
SBS & 11.53 & 35.92 & 102.10 & 315.32\\
\bottomrule
\end{tabular}
}
}
\end{table*}

\subsection{Effectiveness of a pre-solver (RQ3)}
\label{sec:exp_presolver}

Table \ref{tab:effect_presolver} shows results of the pairwise classification-based system with and without a pre-solver.
In Table \ref{tab:effect_presolver}, AS ($50 \times n$) and AS ($100 \times n$) are the systems with the sample size $50 \times n$ and $100 \times n$, respectively.
Recall that the default sample size is $50 \times n$.
SLSQP-AS and SMAC-AS are the system using SLSQP and SMAC as pre-solvers, respectively.
%
%
Tables \ref{tab:effect_presolver}(a) and \ref{tab:effect_presolver}(b) show the median of 31 ``mean relSP1'' values when using $\mathcal{A}_{\mathrm{kt}}$ and $\mathcal{A}_{\mathrm{bmtp}}$ as algorithm portfolios, respectively.
Table \ref{tab:effect_presolver} does not show the results for the RI-CV, but they are similar to the results for the LOIO-CV.
Tables \ref{suptab:effect_presolver_class}--\ref{suptab:effect_presolver_clustering} show results of the five systems with the five portfolios ($\mathcal{A}_{\mathrm{kt}}$, ..., $\mathcal{A}_{\mathrm{mk}}$), respectively.
The symbols $+$ and $-$ indicate that a given system performs significantly better ($+$) and significantly worse ($-$) than AS ($50 \times n$) according to the Wilcoxon rank-sum test with $p < 0.05$.
Apart from the comparison with AS ($50 \times n$), results that are better than those of the SBS are highlighted in \adjustbox{margin=0.1em, bgcolor=c1}{dark gray}.
The comparison with the SBS is discussed later in Section \ref{sec:exp_cv}.

As seen from Table \ref{tab:effect_presolver}, SLSQP-AS performs significantly better than AS ($50 \times n$) for the three cross-validation methods in most cases.
Tables \ref{suptab:effect_presolver_class}--\ref{suptab:effect_presolver_clustering} also show that SLSQP-AS does not perform significantly worse than AS ($50 \times n$) in this study.
These results suggest that the performance of algorithm selection systems can be significantly improved by using SLSQP as a pre-solver.

\begin{figure}[t]
   \centering
\includegraphics[width=0.32\textwidth]{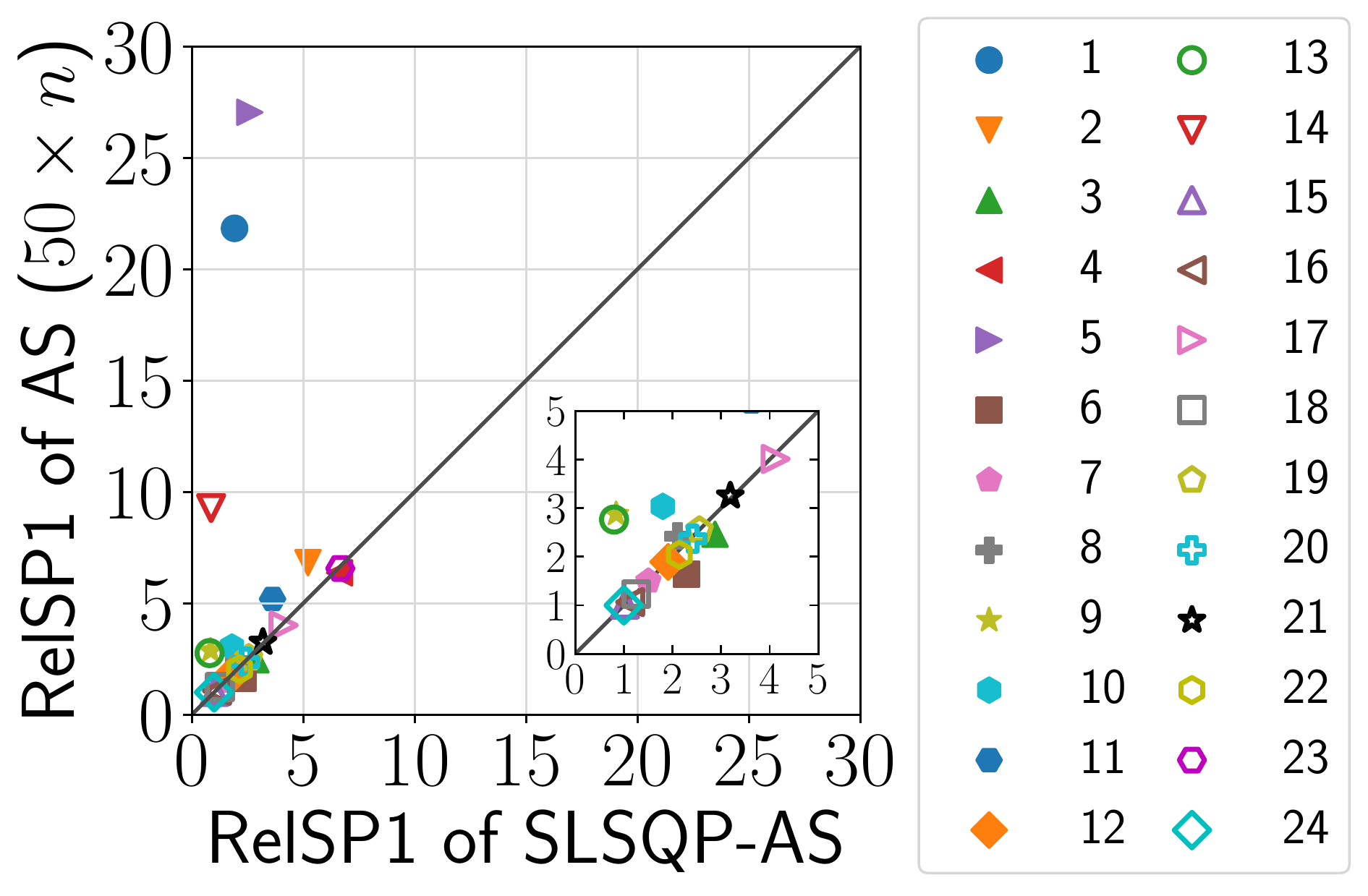}
   \caption{Distribution of 31 relSP1 values of the pairwise classification-based algorithm selection system with $\mathcal{A}_{\mathrm{kt}}$ on the 24 BBOB functions with $n=5$ for the LOIO-CV.}
   \label{fig:comp2}
\end{figure}

Fig. \ref{fig:comp2} shows a comparison of AS ($50 \times n$) and SLSQP-AS on the 24 BBOB functions with $n=5$.
Fig. \ref{fig:comp2} shows the results of a single run with a median ``mean relSP1'' value among 31 runs.
Fig. \ref{supfig:fevals_slsqp} also shows the distribution of the number of function evaluations used by SLSQP in the pre-solving phase.
Recall that the maximum number of function evaluations in the pre-solving phase is $50 \times n$.
As shown in Fig. \ref{fig:comp2}, SLSQP-AS obtains much smaller values of the relSP1 than AS ($50 \times n$) on $f_1$ (the Sphere function), $f_5$ (the Linear Slope function), and $f_{14}$ (the different powers function).
These results demonstrate that the use of SLSQP is highly effective on some unimodal functions.
In other words, $s$ function evaluations can be saved by using SLSQP as the pre-solver.
As noted in \cite{Hansen19}, the poor performance of SLSQP on $f_6$ (the Attractive Sector function) is mainly due to a small number of function evaluations.
%
In contrast, SLSQP does not reach the target value within $50\times n$ function evaluations on the 12 multimodal functions $f_3, f_4, f_{15}, ..., f_{24}$, except for some instances of $f_{21}$ (the Gallagher's Gaussian 101-me Peaks function).
Since the initial $50\times n$ function evaluations in the pre-solving phase are wasted in this case, SLSQP-AS performs slightly worse than AS ($50\times n$) in terms of the relSP1.
However, any optimizer requires a large number of function evaluations for hard multimodal functions to reach the target value.
For this reason, as shown in Fig. \ref{fig:comp2}, a small number of additional function evaluations in the pre-solving phase does not drastically deteriorate the performance of the system.

In contrast to SLSQP-AS, Table \ref{tab:effect_presolver} shows that the performance of SMAC-AS is not better than that of AS ($50 \times n$) in most cases.
Our results suggest that SMAC is not appropriate for a pre-solver.
AS ($100 \times n$) performs significantly worse than AS ($50 \times n$) in most cases, except for the result for $n=2$ in Table \ref{suptab:effect_presolver_class}(h) and the result for $n=3$ in Table \ref{suptab:effect_presolver_pclassfi}(j).
Based on these results, it would be better to allocate half of the budget ($50 \times n$ function evaluations) to the pre-solving and sampling phases, instead of allocating all the budget ($100 \times n$ function evaluations) to the sampling phase.

One may be interested in the influence of the sample size $s$ on the performance of algorithm selection systems.
Table \ref{suptab:effect_sample_size} further shows the comparison of the five algorithm selection system with $s \in \{50 \times n, 25 \times n, 100 \times n, 200 \times n\}$.
$\mathcal{A}_{\mathrm{kt}}$ was used in this comparison.
As shown in Table \ref{suptab:effect_sample_size}, $s \geq 100 \times n$ is less effective than $s = 50 \times n$ in most cases.
In contrast, $s = 25 \times n$ is more effective than $s = 50 \times n$ in some cases.
These results suggest that $s$ can possibly be set to $s \leq 50 \times n$, which is the general setting.
An in-depth analysis of $s$ is another future work.

Let $\mathcal{Y}$ be a set of all solutions found so far by a pre-solver.
As in \cite{HeYL18,JankovicED21}, it is possible to compute features based on the union of the sample $\mathcal{X}$ and $\mathcal{Y}$.
A similar idea was also discussed in \cite{DerbelLVAT19}.
Table \ref{suptab:effect_united_sample} shows a comparison of two SLSQP-AS variants that compute features based on $\mathcal{X}$ and $\mathcal{X} \cup \mathcal{Y}$, respectively.
$\mathcal{A}_{\mathrm{kt}}$ was used in this comparison.
Although SLSQP-AS with $\mathcal{X} \cup \mathcal{Y}$ outperforms SLSQP-AS with $\mathcal{X}$ in some cases, we cannot say that SLSQP-AS with $\mathcal{X} \cup \mathcal{Y}$ generally performs better than SLSQP-AS with $\mathcal{X}$.
Table \ref{suptab:effect_united_sample} also shows that features computed based on $\mathcal{X} \cup \mathcal{Y}$ are less effective than those based on $\mathcal{X}$ alone in some cases.
The unpromising results may be due to the extremely biased distribution of solutions in $\mathcal{Y}$.
Since SLSQP terminates early on some functions, the size of $\mathcal{X} \cup \mathcal{Y}$ is not constant for all the 24 BBOB functions.
However, as revealed in \cite{RenauDDD20}, the sample size should be the same for all the functions to obtain effective features.
Designing a method for ``cleansing'' $\mathcal{Y}$ is a future research topic.

\noindent \colorbox{c2}{\textbf{Answers to RQ3}}
Our results demonstrated that the overall performance of algorithm selection systems can be significantly improved by using SLSQP as a pre-solver in most cases, especially for easy function instances.
%
Although the pre-solving approach has not been considered for black-box optimization, it would be better to incorporate a pre-solver into algorithm selection systems.
We believe that our promising findings here facilitate the use of pre-solvers to researchers in the field of black-box numerical optimization.

\subsection{Comparison of the five algorithm selection methods (RQ4)}
\label{sec:comparison_asmethods}

Based on the results reported in Section \ref{sec:exp_presolver}, we use SLSQP-AS in the rest of this paper.
Tables \ref{tab:aps_5as} and \ref{tab:aps_5as_bmtp} show performance score values of the five algorithm selection systems using $\mathcal{A}_{\mathrm{kt}}$ and $\mathcal{A}_{\mathrm{bmtp}}$, respectively.
Tables \ref{tab:aps_5as} and \ref{tab:aps_5as_bmtp} do not show the results for the RI-CV, which are similar to the results for the LOIO-CV.
The best and second-best results are highlighted in \adjustbox{margin=0.1em, bgcolor=c1}{dark gray} and \adjustbox{margin=0.1em, bgcolor=c2}{gray}, respectively.
Tables \ref{suptab:aps_5as}--\ref{suptab:aps_5as_ls14} show results when using the 14 portfolios.
Tables \ref{suptab:aps_5as_friedman}--\ref{suptab:aps_5as_ls14_friedman} also show results of the Friedman test.

On the one hand, as seen from Table \ref{tab:aps_5as}, for the LOIO-CV and the LOPOAD-CV, the pairwise classification-based system performs the best in most cases when using $\mathcal{A}_{\mathrm{kt}}$.
This observation is consistent with the results for combinatorial optimization shown in \cite{LindauerHHS15}.
For the LOIO-CV, the classification-based system also performs well.
For the LOPO-CV, the regression-based system is the best performer.
The results of the systems with $\mathcal{A}_{\mathrm{dlvat}}$ and $\mathcal{A}_{\mathrm{jped}}$ are similar to those with $\mathcal{A}_{\mathrm{kt}}$.

\begin{table}[t]
\setlength{\tabcolsep}{2.4pt} 
\renewcommand{\arraystretch}{0.95}  
\centering
  \caption{\small Results of the five algorithm selection systems. Tables (a)--(c) show performance score values of the five systems using $\mathcal{A}_{\mathrm{kt}}$ for the three cross-validation methods, respectively.      
}
\label{tab:aps_5as}
{\scriptsize
\subfloat[LOIO-CV]{
\begin{tabular}{lC{1em}C{1em}C{1em}C{1em}}
\toprule
 & \multicolumn{4}{c}{$n$}\\
\cmidrule{2-5}
 & $2$ & $3$ & $5$ & $10$\\
\midrule
Cla. & \cellcolor{c2}1 & \cellcolor{c2}1 & \cellcolor{c1}0 & \cellcolor{c1}0\\
Reg. & 2 & 3 & 2 & 2\\
P-Cla. & \cellcolor{c1}0 & \cellcolor{c1}0 & \cellcolor{c2}1 & \cellcolor{c2}1\\
P-Reg. & 3 & 4 & 3 & 3\\
Clu. & 3 & 2 & 4 & 4\\
\bottomrule
\end{tabular}
}
\subfloat[LOPO-CV]{
\begin{tabular}{lC{1em}C{1em}C{1em}C{1em}}
\toprule
 & \multicolumn{4}{c}{$n$}\\
\cmidrule{2-5}
 & $2$ & $3$ & $5$ & $10$\\
\midrule
Cla. & 3 & 4 & 3 & 3\\
Reg. & \cellcolor{c2}1 & \cellcolor{c1}0 & \cellcolor{c1}0 & \cellcolor{c1}0\\
P-Cla. & \cellcolor{c1}0 & \cellcolor{c2}2 & \cellcolor{c2}1 & 2\\
P-Reg. & 3 & \cellcolor{c1}0 & 2 & \cellcolor{c2}1\\
Clu. & \cellcolor{c1}0 & \cellcolor{c2}2 & 3 & 4\\
\bottomrule
\end{tabular}
}
\subfloat[LOPOAD-CV]{
\begin{tabular}{lC{1em}C{1em}C{1em}C{1em}}
\toprule
 & \multicolumn{4}{c}{$n$}\\
\cmidrule{2-5}
 & $2$ & $3$ & $5$ & $10$\\
\midrule
Cla. & \cellcolor{c2}2 & 4 & 3 & 3\\
Reg. & \cellcolor{c1}0 & \cellcolor{c1}0 & \cellcolor{c2}1 & \cellcolor{c1}0\\
P-Cla. & \cellcolor{c1}0 & \cellcolor{c1}0 & \cellcolor{c1}0 & \cellcolor{c1}0\\
P-Reg. & 4 & \cellcolor{c2}1 & \cellcolor{c2}1 & \cellcolor{c2}1\\
Clu. & \cellcolor{c1}0 & \cellcolor{c1}0 & 3 & 4\\
\bottomrule
\end{tabular}
}
}

\centering
  \caption{\small Results of the five systems using $\mathcal{A}_{\mathrm{bmtp}}$.
}
\label{tab:aps_5as_bmtp}
{\scriptsize
\subfloat[LOIO-CV]{
\begin{tabular}{lC{1em}C{1em}C{1em}C{1em}}
\toprule
 & \multicolumn{4}{c}{$n$}\\
\cmidrule{2-5}
 & $2$ & $3$ & $5$ & $10$\\
\midrule
Cla. & \cellcolor{c1}0 & 3 & \cellcolor{c1}0 & \cellcolor{c2}2\\
Reg. & \cellcolor{c2}3 & \cellcolor{c2}1 & 2 & \cellcolor{c1}0\\
P-Cla. & \cellcolor{c1}0 & \cellcolor{c1}0 & \cellcolor{c2}1 & \cellcolor{c1}0\\
P-Reg. & \cellcolor{c1}0 & \cellcolor{c1}0 & \cellcolor{c2}1 & \cellcolor{c1}0\\
Clu. & 4 & 4 & 4 & 4\\
\bottomrule
\end{tabular}
}
\subfloat[LOPO-CV]{
\begin{tabular}{lC{1em}C{1em}C{1em}C{1em}}
\toprule
 & \multicolumn{4}{c}{$n$}\\
\cmidrule{2-5}
 & $2$ & $3$ & $5$ & $10$\\
\midrule
Cla. & \cellcolor{c1}0 & 2 & \cellcolor{c2}3 & 3\\
Reg. & \cellcolor{c1}0 & \cellcolor{c1}0 & \cellcolor{c1}0 & \cellcolor{c1}0\\
P-Cla. & \cellcolor{c1}0 & 2 & \cellcolor{c1}0 & \cellcolor{c2}2\\
P-Reg. & \cellcolor{c2}1 & \cellcolor{c2}1 & \cellcolor{c1}0 & \cellcolor{c1}0\\
Clu. & 4 & 2 & 4 & 4\\
\bottomrule
\end{tabular}
}
\subfloat[LOPOAD-CV]{
\begin{tabular}{lC{1em}C{1em}C{1em}C{1em}}
\toprule
 & \multicolumn{4}{c}{$n$}\\
\cmidrule{2-5}
 & $2$ & $3$ & $5$ & $10$\\
\midrule
Cla. & \cellcolor{c2}3 & \cellcolor{c1}0 & \cellcolor{c1}0 & \cellcolor{c1}0\\
Reg. & \cellcolor{c1}0 & 3 & 3 & 3\\
P-Cla. & \cellcolor{c1}0 & \cellcolor{c1}0 & 2 & \cellcolor{c2}1\\
P-Reg. & \cellcolor{c1}0 & \cellcolor{c2}2 & \cellcolor{c2}1 & 2\\
Clu. & 4 & 4 & 4 & 4\\
\bottomrule
\end{tabular}
}
}
\end{table}

On the other hand, as seen from Table \ref{tab:aps_5as_bmtp}, for the LOIO-CV, the pairwise regression-based and pairwise classification-based systems have the same performance when using $\mathcal{A}_{\mathrm{bmtp}}$.
For the LOPO-CV, the pairwise regression-based system is competitive with the regression-based system  for $n \in \{5, 10\}$.
These results indicate that the best algorithm selection method depends on algorithm portfolios and cross-validation methods.

Since $|\mathcal{A}_{\mathrm{kt}}|=12$ and $|\mathcal{A}_{\mathrm{bmtp}}|=4$, one may think that the portfolio size determines the best algorithm selection method.
To investigate the scalability of the five systems with respect to the portfolio size, Tables \ref{suptab:aps_5as_ls2} and \ref{suptab:aps_5as_ls14} show results of the five systems with $\mathcal{A}_{\mathrm{ls}2}$, ...,  $\mathcal{A}_{\mathrm{ls}18}$, where $|\mathcal{A}_{\mathrm{ls}2}| = 2$, ..., $|\mathcal{A}_{\mathrm{ls}18}|=18$.
As seen from Tables \ref{suptab:aps_5as_ls2} and \ref{suptab:aps_5as_ls14}, there is no clear correlation between the portfolio size and the performance rankings of the five systems.
For example, for the LOIO-CV, the pairwise regression-based system does not obtain the best performance on any dimension when using $\mathcal{A}_{\mathrm{ls}4}$ (Table \ref{suptab:aps_5as_ls2}(d)).
In contrast, for the LOPO-CV, the pairwise regression-based system performs the best for all dimensions when using $\mathcal{A}_{\mathrm{ls}12}$ (Table \ref{suptab:aps_5as_ls2}(q)).
Note that $|\mathcal{A}_{\mathrm{bmtp}}| = |\mathcal{A}_{\mathrm{ls}4}|$ and $|\mathcal{A}_{\mathrm{kt}}| = |\mathcal{A}_{\mathrm{ls}12}|$.
These results suggest that components of portfolios are more important than the portfolio size to determine the best algorithm selection method.
Such an observation has not been reported even in combinatorial optimization.

One exception is the results of the classification-based system for the LOPOAD-CV.
Tables \ref{tab:aps_5as}, \ref{tab:aps_5as_bmtp}, \ref{suptab:aps_5as}--\ref{suptab:aps_5as_ls14} show that the classification-based system is the best performer for some dimensions when using a small-size portfolio (e.g., $\mathcal{A}_{\mathrm{bmtp}}$ and $\mathcal{A}_{\mathrm{ls}6}$)
In contrast, the classification-based system cannot perform the best when using a large-size portfolio (e.g., $\mathcal{A}_{\mathrm{kt}}$, $\mathcal{A}_{\mathrm{ls}10}$, ..., $\mathcal{A}_{\mathrm{ls}18}$).
The poor scalability of the classification-based system is simply because multi-class classification with many classes is difficult.
Here, the classification-based system uses a random forest model (see Section \ref{sec:selection_methods}).



\noindent \colorbox{c2}{\textbf{Answers to RQ4}}
Our results showed that the best algorithm selection method mainly depends on the components of algorithm portfolios.
For the challenging LOPO-CV, the regression-based method is the best performer, followed by the pairwise regression-based method.
This observation suggests that the regression-based method is likely to perform the best when the training and testing instances are quite dissimilar as in the LOPO-CV.


\begin{figure}[t]
   \centering
\subfloat[Pairwise classification]{  
\includegraphics[width=0.3\textwidth]{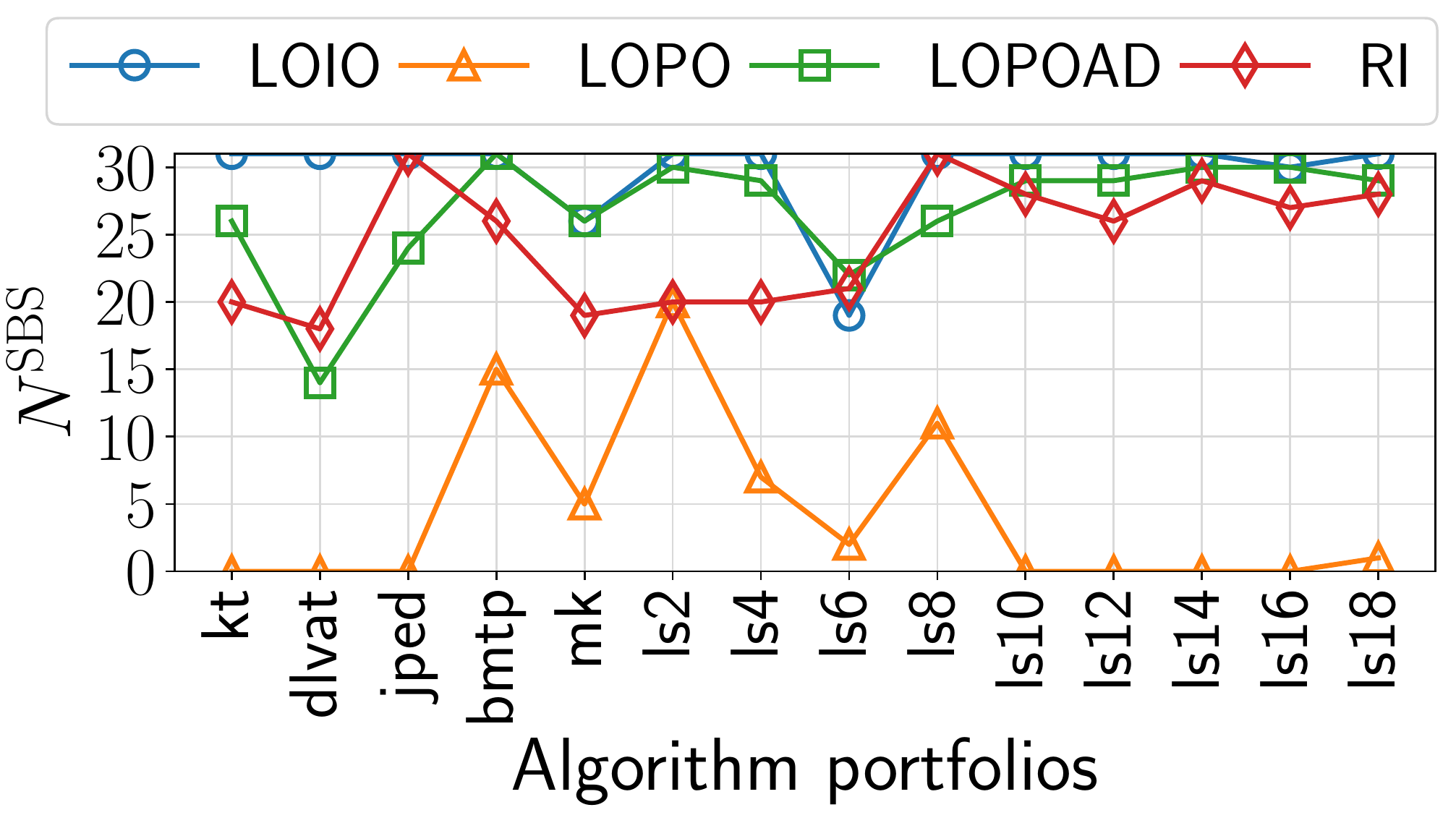}
}
\\
\subfloat[Regression]{  
\includegraphics[width=0.3\textwidth]{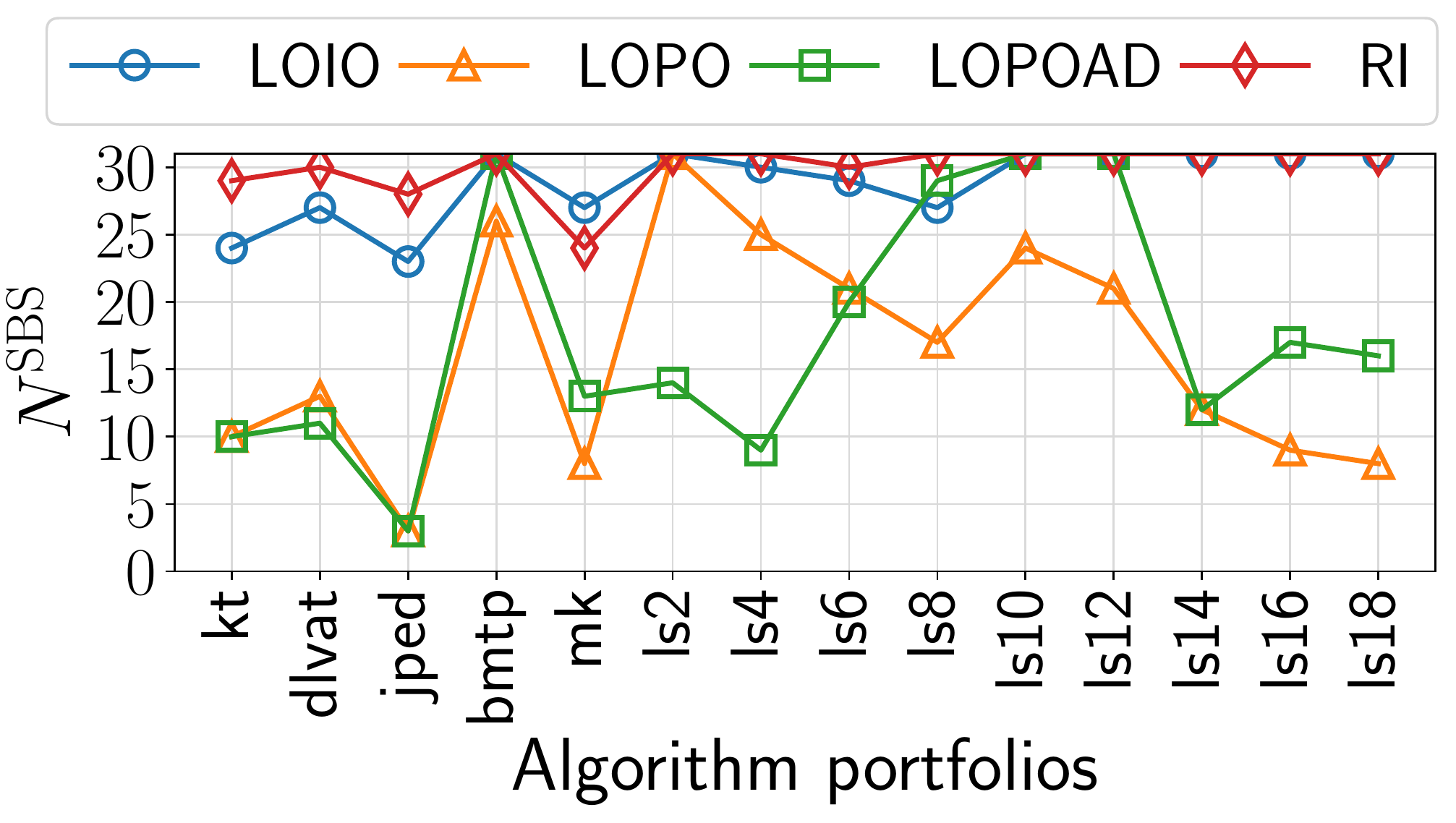}
}
\caption{$N^{\mathrm{SBS}}$ in which the pairwise classification-based and regression-based systems outperform the SBS for $n=5$.}
   \label{fig:n_beat_sbs}
\end{figure}

\subsection{On the difficulty of outperforming the SBS (RQ5)}
\label{sec:exp_cv}

Fig. \ref{fig:n_beat_sbs} shows a comparison of the SBS and the pairwise classification-based and regression-based systems according to their 31 mean relSP1 values for $n=5$.
Fig. \ref{fig:n_beat_sbs} shows the number of times ($N^{\mathrm{SBS}} \in [0, 31]$) in which a system outperforms the SBS over 31 runs.
We use the 14 portfolios.
We measure the difficulty of outperforming the SBS based on $N^{\mathrm{SBS}}$.
If a system with a portfolio $\mathcal{A}$ achieves a small $N^{\mathrm{SBS}}$ value, we say that  it is difficult for the system to outperform the SBS in $\mathcal{A}$.
Figs. \ref{fig:n_beat_sbs_c}--\ref{fig:n_beat_sbs_cl} show results of all the five systems for $n \in \{2, 3, 5, 10\}$.

As shown in Fig. \ref{fig:n_beat_sbs}, for the LOIO-CV and the LOPOAD-CV, the pairwise classification-based system outperforms the regression-based system for most portfolios in terms of $N^{\mathrm{SBS}}$.
In contrast, for the LOPO-CV and the RI-CV, the regression-based system achieves a better $N^{\mathrm{SBS}}$ value than the pairwise classification-based system for all the 14 portfolios.
These results based on $N^{\mathrm{SBS}}$ are consistent with the results shown in Section \ref{sec:comparison_asmethods}.
This is the first study to show that a system outperforms the SBS for the LOIO-CV and the LOPO-CV for black-box numerical optimization \textit{when considering the number of function evaluations used in the sampling phase}.


%
%
As seen from Fig. \ref{fig:n_beat_sbs}, $N^{\mathrm{SBS}}$ depends on portfolios and cross-validation methods.
However, we cannot find the correlation between the portfolio size and $N^{\mathrm{SBS}}$.
For the LOIO-CV, Fig. \ref{fig:n_beat_sbs}(a) shows that the pairwise classification-based system performs better than the SBS for all 31 runs when using 11 out of the 14 portfolios.
The results for the LOPOAD-CV, the RI-CV, and the LOIO-CV are similar.
For the LOPO-CV, when using 7 out of the 14 portfolios, the pairwise classification-based system obtains $N^{\mathrm{SBS}}=0$.
Our results show that it is generally the most difficult to outperform the SBS for the LOPO-CV.
Of course, this is not always true.
For example, Fig. \ref{fig:n_beat_sbs}(b) shows that $N^{\mathrm{SBS}}$ obtained by the regression-based system for the LOPO-CV is larger than that for the LOPOAD-CV when using $\mathcal{A}_{\mathrm{dlvat}}$, $\mathcal{A}_{\mathrm{ls}2}$, $\mathcal{A}_{\mathrm{ls}4}$, and $\mathcal{A}_{\mathrm{ls}6}$.

Although the above discussions are based on the results for $n=5$, Figs. \ref{fig:n_beat_sbs_c}--\ref{fig:n_beat_sbs_cl} show that $N^{\mathrm{SBS}}$ depends on the dimension $n$.
As shown in Figs. \ref{fig:n_beat_sbs_c}--\ref{fig:n_beat_sbs_cl}, $N^{\mathrm{SBS}}$ for $n \in \{2, 3\}$ is generally smaller than $N^{\mathrm{SBS}}$ for $n \in \{5, 10\}$.
This is because the SBS can reach the target value for a relatively small number of function evaluations on low-dimensional function instances, even including hard multimodal instances.
In this case, the budget of function evaluations used in the sampling phase is a critical disadvantage for algorithm selection.

Below, we discuss which cross-validation method should be used for benchmarking algorithm selection systems.
In \cite{vanSteinWB20}, van Stein et al. analyzed the fitness landscape of a neural architecture search (NAS) problem with $n=23$.
Their results based on the ELA approach revealed that the fitness landscape of the NAS problem is different from that of any BBOB function.
We do not intend to generalize their conclusion, but it is practical to assume that problem instances in the testing and training phases are different.
The LOPO-CV is appropriate for this purpose.
%
%
In contrast to the LOPO-CV, it would \textit{not} be better to use the LOIO-CV.
In the LOIO-CV, function instances used in the training and testing phases are always similar.
If a researcher uses the LOIO-CV, she/he can overestimate the performance of an algorithm selection system that does not actually work well for any real-world setting.
For a similar reason, we do not suggest using the RI-CV.


Let $f^{\mathrm{real}}$ with $n^{\mathrm{real}}$ be an $n^{\mathrm{real}}$-dimensional real-world problem.
It is rare that $f^{\mathrm{real}}$ instances with $n \neq n^{\mathrm{real}}$ are available in the training phase.
It is also practical to use the same $n$ in the training and testing phases.
Our results also showed that the LOPOAD-CV is less challenging than the LOPO-CV.
It may be better not to use the LOPOAD-CV without a particular reason.

\noindent \colorbox{c2}{\textbf{Answers to RQ5}}
We demonstrated that the difficulty of outperforming the SBS depends on 
algorithm portfolios and dimensions.
For example, even for the LOPO-CV, our results showed that the regression-based system can often outperform the SBS when using $\mathcal{A}_{\mathrm{bmtp}}$.
Since a result obtained using a single portfolio can be misleading, it would be better to use multiple portfolios (e.g., $\mathcal{A}_{\mathrm{bmtp}}$ and $\mathcal{A}_{\mathrm{kt}}$) for benchmarking algorithm selection systems.
Based on the discussion, we suggest using the LOPO-CV.



\subsection{Comparison of algorithm portfolios (RQ6)}
\label{sec:exp_ap}

Table \ref{tab:vbs_sbs} shows the mean relSP1 values of the VBS and the SBS in the 14 portfolios.
While the other sections calculate the relSP1 value based on each portfolio, only this section calculates the relSP1 value based on the union of all the 14 portfolios $\mathcal{A}_{\mathrm{kt}} \cup \cdots \cup \mathcal{A}_{\mathrm{ls}18}$.
For each function, the best SP1 value for the relSP1 calculation is obtained from the results of all optimizers in the 14 portfolios (see Table \ref{tab:ap_ls}).
For this reason, the relSP1 value of even the VBS is not 1.

As shown in Table \ref{tab:vbs_sbs}, a larger-size portfolio achieves better VBS performance.
While the VBS performance of $\mathcal{A}_{\mathrm{ls}18}$ is the best for any $n$, that in $\mathcal{A}_{\mathrm{bmtp}}$ and $\mathcal{A}_{\mathrm{mk}}$ is the worst for $n \in \{2, 3\}$ and $n \in \{5, 10\}$, respectively.
For $n \in \{5, 10\}$, HCMA is the SBS in $\mathcal{A}_{\mathrm{kt}}$, $\mathcal{A}_{\mathrm{dlvat}}$, $\mathcal{A}_{\mathrm{jped}}$, $\mathcal{A}_{\mathrm{ls}2}$, $\mathcal{A}_{\mathrm{ls}4}$, and $\mathcal{A}_{\mathrm{ls}6}$.
HCMA is also the best optimizer in the union of the 14 portfolios.
Since $\mathcal{A}_{\mathrm{bmtp}}$, $\mathcal{A}_{\mathrm{mk}}$, $\mathcal{A}_{\mathrm{ls}8}$, ..., $\mathcal{A}_{\mathrm{ls}18}$ do not include HCMA, their SBS performance is poor for $n \in \{5, 10\}$.
These results indicate that optimizing the VBS performance of $\mathcal{A}$ does not always mean optimizing the SBS performance of $\mathcal{A}$.

Table \ref{tab:aps_14ap} shows performance score values of the pairwise classification-based system with the 14 portfolios for the three cross-validation methods.
We do not show the results for the RI-CV, which are similar to the results for the LOIO-CV.
As in Tables \ref{tab:aps_5as} and \ref{tab:aps_5as_bmtp}, the best and second-best results are highlighted in \adjustbox{margin=0.1em, bgcolor=c1}{dark gray} and \adjustbox{margin=0.1em, bgcolor=c2}{gray}, respectively.
Tables \ref{suptab:aps_14ap_cla}--\ref{suptab:aps_14ap_clu} show the results of the five systems, respectively.
With some exceptions, the results of the other four systems are relatively similar to the results in Table \ref{tab:aps_14ap}.
Tables \ref{suptab:aps_14ap_cla_friedman}--\ref{suptab:aps_14ap_clu_friedman} also show results of the Friedman test.

\begin{table}[t]
\setlength{\tabcolsep}{2.4pt} 
\renewcommand{\arraystretch}{0.1}  
\centering
  \caption{\small Mean relSP1 values of the VBS and the SBS in the 14 algorithm portfolios.
    }
\label{tab:vbs_sbs}
{\scriptsize
  \begin{tabular}{lcccc|ccccccccc}
  \toprule
AP & \multicolumn{4}{c}{VBS} & \multicolumn{4}{c}{SBS}\\
\cmidrule(lr){2-5} \cmidrule(lr){6-9}
& $n=2$ & $n=3$ & $n=5$ & $n=10$ & $n=2$ & $n=3$ & $n=5$ & $n=10$\\
\midrule
$\mathcal{A}_{\mathrm{kt}}$  & 2.59 & 1.76 & 1.62 & 1.56 & 15.71 & 14.36 & 6.63 & 9.53\\
\midrule
$\mathcal{A}_{\mathrm{dlvat}}$  & 3.19 & 1.84 & 1.64 & 1.60 & 15.71 & 14.36 & 6.63 & 9.53\\
\midrule
$\mathcal{A}_{\mathrm{jped}}$  & 2.52 & 1.83 & 1.52 & 1.58 & 15.71 & 14.36 & 6.63 & 9.53\\
\midrule
$\mathcal{A}_{\mathrm{bmtp}}$  & 75.22 & 56.16 & 5.20 & 5.55 & 115.88 & 174.38 & 40173.96 & 46560.09\\
\midrule
$\mathcal{A}_{\mathrm{mk}}$  & 14.96 & 54.71 & 6.77 & 8.63 & 33.06 & 174.38 & 40173.96 & 46560.09\\
\midrule
$\mathcal{A}_{\mathrm{ls}2}$  & 9.58 & 7.04 & 3.37 & 2.37 & 15.71 & 14.36 & 6.63 & 9.53\\
\midrule
$\mathcal{A}_{\mathrm{ls}4}$  & 2.84 & 5.75 & 2.60 & 1.71 & 15.71 & 14.36 & 6.63 & 9.53\\
\midrule
$\mathcal{A}_{\mathrm{ls}6}$  & 2.43 & 4.65 & 1.54 & 1.43 & 15.71 & 14.36 & 6.63 & 9.53\\
\midrule
$\mathcal{A}_{\mathrm{ls}8}$  & 2.17 & 4.61 & 1.24 & 1.67 & 15.71 & 14.36 & 40205.16 & 23745.28\\
\midrule
$\mathcal{A}_{\mathrm{ls}10}$  & 2.12 & 4.49 & 1.20 & 1.12 & 15.71 & 14.36 & 40205.16 & 532.13\\
\midrule
$\mathcal{A}_{\mathrm{ls}12}$  & 2.10 & 4.20 & 1.12 & 1.09 & 15.71 & 14.36 & 40205.16 & 532.13\\
\midrule
$\mathcal{A}_{\mathrm{ls}14}$  & 2.00 & 1.21 & 1.13 & 1.11 & 15.71 & 14.36 & 40156.68 & 532.13\\
\midrule
$\mathcal{A}_{\mathrm{ls}16}$  & 1.04 & 1.20 & 1.03 & 1.10 & 27.63 & 134.12 & 40156.68 & 532.13\\
\midrule
$\mathcal{A}_{\mathrm{ls}18}$  & 1.02 & 1.18 & 1.02 & 1.06 & 27.63 & 134.12 & 40156.68 & 532.13\\
\bottomrule
  \end{tabular}
}
 \end{table}
 \begin{table}[t]
 \setlength{\tabcolsep}{2.4pt} 
\renewcommand{\arraystretch}{0.85}  
\centering
  \caption{\small Results of the pairwise classification-based algorithm selection systems with the 14 algorithm portfolios for $n \in \{2, 3, 5, 10\}$. Tables (a)--(c) show the performance score values for the three cross-validation methods, respectively.    
}
\label{tab:aps_14ap}
{\scriptsize
\subfloat[LOIO-CV]{
\begin{tabular}{lC{1em}C{1em}C{1em}C{1em}}
\toprule
 & \multicolumn{4}{c}{$n$}\\
\cmidrule{2-5}
 & $2$ & $3$ & $5$ & $10$\\
\midrule
$\mathcal{A}_{\mathrm{kt}}$ & \cellcolor{c1}0 & \cellcolor{c1}0 & \cellcolor{c1}0 & \cellcolor{c1}0\\
$\mathcal{A}_{\mathrm{dlvat}}$ & \cellcolor{c1}0 & \cellcolor{c1}0 & \cellcolor{c1}0 & \cellcolor{c1}0\\
$\mathcal{A}_{\mathrm{jped}}$ & \cellcolor{c1}0 & \cellcolor{c1}0 & \cellcolor{c1}0 & \cellcolor{c1}0\\
$\mathcal{A}_{\mathrm{bmtp}}$ & 8 & 7 & 12 & 6\\
$\mathcal{A}_{\mathrm{mk}}$ & 6 & 6 & 12 & 6\\
$\mathcal{A}_{\mathrm{ls}2}$ & 9 & 5 & \cellcolor{c2}3 & \cellcolor{c2}4\\
$\mathcal{A}_{\mathrm{ls}4}$ & 9 & 7 & 4 & \cellcolor{c1}0\\
$\mathcal{A}_{\mathrm{ls}6}$ & 11 & 9 & 5 & \cellcolor{c2}4\\
$\mathcal{A}_{\mathrm{ls}8}$ & \cellcolor{c2}4 & \cellcolor{c2}4 & 11 & 6\\
$\mathcal{A}_{\mathrm{ls}10}$ & \cellcolor{c2}4 & \cellcolor{c2}4 & 6 & 8\\
$\mathcal{A}_{\mathrm{ls}12}$ & 6 & \cellcolor{c2}4 & 6 & 8\\
$\mathcal{A}_{\mathrm{ls}14}$ & \cellcolor{c1}0 & \cellcolor{c1}0 & 6 & 8\\
$\mathcal{A}_{\mathrm{ls}16}$ & 12 & 11 & 6 & 8\\
$\mathcal{A}_{\mathrm{ls}18}$ & 12 & 11 & 6 & 8\\
\bottomrule
\end{tabular}
}
\subfloat[LOPO-CV]{
\begin{tabular}{lC{1em}C{1em}C{1em}C{1em}}
\toprule
 & \multicolumn{4}{c}{$n$}\\
\cmidrule{2-5}
 & $2$ & $3$ & $5$ & $10$\\
\midrule
$\mathcal{A}_{\mathrm{kt}}$ & \cellcolor{c1}0 & 5 & \cellcolor{c2}1 & 3\\
$\mathcal{A}_{\mathrm{dlvat}}$ & \cellcolor{c2}1 & \cellcolor{c1}0 & 5 & 3\\
$\mathcal{A}_{\mathrm{jped}}$ & \cellcolor{c1}0 & 3 & \cellcolor{c2}1 & 3\\
$\mathcal{A}_{\mathrm{bmtp}}$ & 8 & 11 & 5 & 6\\
$\mathcal{A}_{\mathrm{mk}}$ & \cellcolor{c2}1 & 4 & 9 & 7\\
$\mathcal{A}_{\mathrm{ls}2}$ & \cellcolor{c1}0 & \cellcolor{c1}0 & \cellcolor{c1}0 & \cellcolor{c1}0\\
$\mathcal{A}_{\mathrm{ls}4}$ & \cellcolor{c1}0 & \cellcolor{c2}1 & \cellcolor{c1}0 & \cellcolor{c2}1\\
$\mathcal{A}_{\mathrm{ls}6}$ & \cellcolor{c1}0 & 5 & \cellcolor{c2}1 & \cellcolor{c2}1\\
$\mathcal{A}_{\mathrm{ls}8}$ & 5 & 5 & 5 & 8\\
$\mathcal{A}_{\mathrm{ls}10}$ & 10 & 10 & 12 & 7\\
$\mathcal{A}_{\mathrm{ls}12}$ & 10 & 8 & 12 & 7\\
$\mathcal{A}_{\mathrm{ls}14}$ & 5 & 9 & 6 & 7\\
$\mathcal{A}_{\mathrm{ls}16}$ & 10 & 10 & 10 & 7\\
$\mathcal{A}_{\mathrm{ls}18}$ & 9 & 10 & 6 & 7\\
\bottomrule
\end{tabular}
}
\subfloat[LOPOAD-CV]{
\begin{tabular}{lC{1em}C{1em}C{1em}C{1em}}
\toprule
 & \multicolumn{4}{c}{$n$}\\
\cmidrule{2-5}
 & $2$ & $3$ & $5$ & $10$\\
\midrule
$\mathcal{A}_{\mathrm{kt}}$ & \cellcolor{c1}0 & \cellcolor{c2}1 & \cellcolor{c1}0 & \cellcolor{c2}1\\
$\mathcal{A}_{\mathrm{dlvat}}$ & \cellcolor{c2}1 & \cellcolor{c2}1 & 6 & \cellcolor{c2}1\\
$\mathcal{A}_{\mathrm{jped}}$ & \cellcolor{c1}0 & \cellcolor{c2}1 & \cellcolor{c1}0 & \cellcolor{c2}1\\
$\mathcal{A}_{\mathrm{bmtp}}$ & 10 & \cellcolor{c1}0 & 7 & 6\\
$\mathcal{A}_{\mathrm{mk}}$ & 8 & 5 & 11 & 7\\
$\mathcal{A}_{\mathrm{ls}2}$ & 2 & 5 & \cellcolor{c1}0 & \cellcolor{c1}0\\
$\mathcal{A}_{\mathrm{ls}4}$ & 3 & 8 & \cellcolor{c1}0 & \cellcolor{c2}1\\
$\mathcal{A}_{\mathrm{ls}6}$ & \cellcolor{c1}0 & 7 & \cellcolor{c1}0 & 5\\
$\mathcal{A}_{\mathrm{ls}8}$ & 2 & \cellcolor{c2}1 & 6 & 13\\
$\mathcal{A}_{\mathrm{ls}10}$ & 2 & \cellcolor{c2}1 & \cellcolor{c2}5 & 8\\
$\mathcal{A}_{\mathrm{ls}12}$ & 2 & \cellcolor{c2}1 & \cellcolor{c2}5 & 8\\
$\mathcal{A}_{\mathrm{ls}14}$ & \cellcolor{c2}1 & 7 & \cellcolor{c2}5 & 8\\
$\mathcal{A}_{\mathrm{ls}16}$ & 10 & 12 & 7 & 8\\
$\mathcal{A}_{\mathrm{ls}18}$ & 10 & 11 & 7 & 8\\
\bottomrule
\end{tabular}
}
}
\end{table}

As seen from Table \ref{tab:aps_14ap}, the effectiveness of portfolios depends on cross-validation methods.
For the LOIO-CV, the system with the three variants of $\mathcal{A}_{\mathrm{kt}}$ ($\mathcal{A}_{\mathrm{kt}}$, $\mathcal{A}_{\mathrm{dlvat}}$, and $\mathcal{A}_{\mathrm{jped}}$) performs the best for any $n$.
For the LOPO-CV, $\mathcal{A}_{\mathrm{ls}2}$ and $\mathcal{A}_{\mathrm{ls}4}$ are the most effective for the system.
For the LOPOAD-CV, the system performs well when using the three variants of $\mathcal{A}_{\mathrm{kt}}$.
In addition, $\mathcal{A}_{\mathrm{ls}2}$ and $\mathcal{A}_{\mathrm{ls}4}$ are still effective for $n \in \{5, 10\}$.
These results suggest that a small-size portfolio including HCMA is effective when function instances in the training and testing phases are significantly different.
Otherwise, the variants of $\mathcal{A}_{\mathrm{kt}}$ are appropriate for algorithm selection systems.
Based on the poor performance of the system with $\mathcal{A}_{\mathrm{ls}16}$ and $\mathcal{A}_{\mathrm{ls}18}$, we can say that a portfolio of size less than 16 may be effective.
Although a general rule of thumb ``set the portfolio size as small as possible'' has been accepted in the literature, this is the first study to show the correctness of the rule.

Although the VBS performance of $\mathcal{A}_{\mathrm{bmtp}}$ and $\mathcal{A}_{\mathrm{mk}}$ is worst, Table \ref{tab:aps_14ap} shows that the system with $\mathcal{A}_{\mathrm{bmtp}}$ and $\mathcal{A}_{\mathrm{mk}}$ achieves better performance score values than that with $\mathcal{A}_{\mathrm{ls}16}$ and $\mathcal{A}_{\mathrm{ls}18}$ in some cases.
Our results also show that the system with $\mathcal{A}_{\mathrm{bmtp}}$ and $\mathcal{A}_{\mathrm{mk}}$ performs the best in a few cases, e.g., the result for $n=3$ in Table \ref{tab:aps_14ap}(c).
%
Since the VBS performance of $\mathcal{A}_{\mathrm{ls}12}$ is better than that of $\mathcal{A}_{\mathrm{kt}}$ except for $n=3$, $\mathcal{A}_{\mathrm{ls}12}$ is likely more effective than $\mathcal{A}_{\mathrm{kt}}$.
Unexpectedly, Table \ref{tab:aps_14ap} and Tables \ref{suptab:aps_14ap_cla}--\ref{suptab:aps_14ap_clu} show that $\mathcal{A}_{\mathrm{kt}}$ is more effective than $\mathcal{A}_{\mathrm{ls}12}$ except for a few cases.
%
These results suggest that the VBS performance of a portfolio $\mathcal{A}$ does not always represent the effectiveness of $\mathcal{A}$.
This is due to the difficulty of selecting the best algorithm, especially for the LOPO-CV.
Our observation can be a useful clue to construct effective algorithm portfolios.

\noindent \colorbox{c2}{\textbf{Answers to RQ6}}
%
%
We found that a small-size portfolio (i.e., $\mathcal{A}_{\mathrm{ls}2}$ and $\mathcal{A}_{\mathrm{ls}4}$) is generally the most effective for the LOPO-CV.
Our results showed that the effectiveness of an algorithm portfolio $\mathcal{A}$ depends not only on its size $|\mathcal{A}|$ but also on its components.
We also demonstrated that a portfolio $\mathcal{A}$ with high VBS performance is not always effective.


\section{Conclusion}
\label{sec:conclusion}

We have investigated the performance of algorithm selection systems for black-box numerical optimization.
Through a benchmarking study, we have answered the six research questions (\textbf{RQ1}--\textbf{RQ6}).
Our findings can contribute to the design of more efficient algorithm selection systems.
For example, we showed that using SLSQP as a pre-solver can significantly improve the performance of algorithm selection systems.
We found that the regression-based selection method performs well for the practical LOPO-CV.
We also demonstrated that a small-size portfolio is generally effective for the LOPO-CV.




As in previous studies \cite{BischlMTP12,AbellMT13,DerbelLVAT19,KerschkeT19,JankovicPED21,MunozK21}, we fixed the target value $f_{\mathrm{target}}$ and the number of functions and instances.
We focused only on functions with up to $n=10$.
We also focused only on the fixed-target scenario.
In addition, there is room for analysis of algorithm selection for another type of black-box optimization, e.g., constrained black-box optimization.
There is much room for investigation of the algorithm portfolio construction for real-world black-box numerical optimization.
An investigation of these factors is need in future research.
An analysis of the performance of algorithm selection systems on real-world applications is also another topic for future work.


We believe that our findings contribute to the standardization of a benchmarking methodology for black-box optimization.
Our results showed that the best algorithm selection system depends on various factors.
Since hand-tuning is difficult in practice, automatic configuration of algorithm selection systems as in \textsc{AutoFolio} \cite{LindauerHHS15} is promising.
It is also interesting to compare feature-based offline algorithm selection systems with online ones (e.g., \cite{BaudisP14}) and rule-based ones (e.g., \cite{LiuMPRRTT20}) in the same platform.
\section*{Acknowledgment}

This work was supported by Leading Initiative for Excellent Young Researchers, MEXT, Japan.




\ifCLASSOPTIONcaptionsoff
  \newpage
\fi



%



\bibliography{reference}
\bibliographystyle{IEEEtran}

%








\begin{IEEEbiography}[{\includegraphics[width=1in,height=1.25in,keepaspectratio]{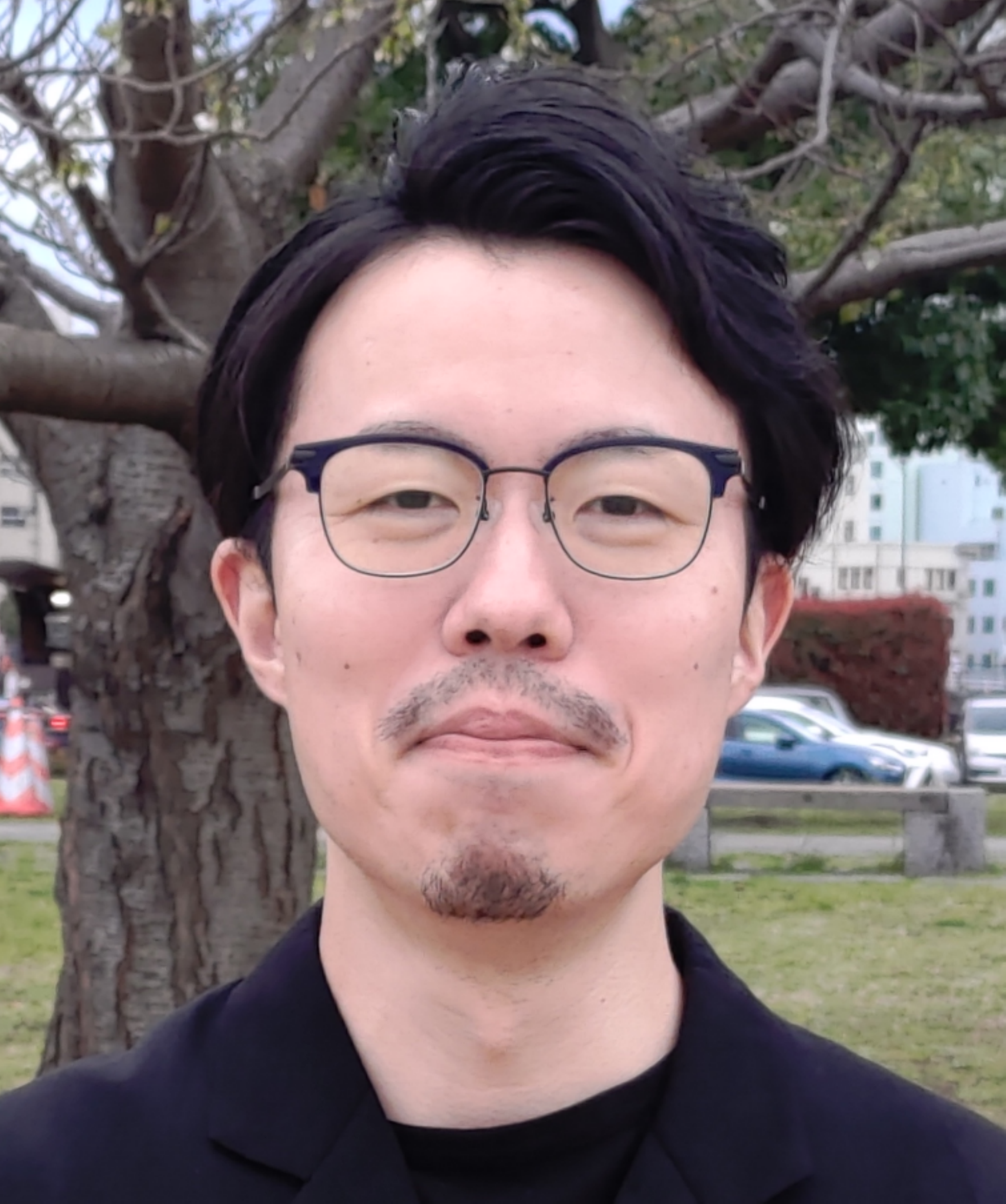}}]{Ryoji Tanabe}
is an Assistant Professor at Yokohama National University, Japan (2019--).
Previously, he was a Research Assistant Professor at Southern University of Science and Technology, China (2017--2019).
He was also a Post-Doctoral Researcher at Japan Aerospace Exploration Agency, Japan (2016--2017).
He received the Ph.D. degree in Science from The University of Tokyo, Japan, in 2016.
His research interests include single- and multi-objective black-box optimization, analysis of evolutionary algorithms, and automatic algorithm configuration.
  \end{IEEEbiography}

\clearpage
\appendix

\input{supplement.tex}

\end{document}